\documentclass[pdflatex,sn-mathphys-num]{sn-jnl}

\usepackage{graphicx}%
\usepackage{multirow}%
\usepackage{amsmath,amssymb,amsfonts}%
\usepackage{amsthm}%
\usepackage{mathrsfs}%
\usepackage[title]{appendix}%
\usepackage{xcolor}%
\usepackage{textcomp}%
\usepackage{manyfoot}%
\usepackage{booktabs}%
\usepackage{algorithm}%
\usepackage{algorithmicx}%
\usepackage{algpseudocode}%
\usepackage{listings}%

\usepackage{eso-pic}
\usepackage{xcolor}

\newcommand{\underreviewstamp}{
  \AddToShipoutPictureFG{
    \AtPageUpperLeft{
      \raisebox{-1.0cm}{
        \hspace{1.0cm}
        {\bfseries Preprint}
      }
    }
  }
}

\usepackage[table]{xcolor}
\usepackage{wasysym}

\theoremstyle{thmstyleone}%
\theoremstyle{thmstyletwo}%

\theoremstyle{thmstylethree}%

\begin{document}

\underreviewstamp

\title[Dual Prototype-Conditioned Diffusion for MUAD]{Dual Prototype-Conditioned Diffusion Model for Scalable Multi-Class Unsupervised Anomaly Detection in Large Category Spaces}


\author[1]{\fnm{Yaoxuan} \sur{Feng}}
\author[1]{\fnm{Yuxin} \sur{Li}}
\author[1]{\fnm{Weijiang} \sur{Lv}}
\author[1]{\fnm{Zixuan} \sur{Zhao}}
\author[1]{\fnm{Yubiao} \sur{Wang}}
\author*[1]{\fnm{Wenchao} \sur{Chen}}\email{chenwenchao@xidian.edu.cn} 
\author*[1]{\fnm{Bo} \sur{Chen}}\email{bchen@mail.xidian.edu.cn} 
\author[1]{\fnm{Hongwei} \sur{Liu}}

\affil*[1]{\orgname{National Key Laboratory of Radar Signal Processing, Xidian University},
\orgaddress{\city{Xi'an}, \postcode{710071}, \country{China}}}


\abstract{
Multi-class anomaly detection aims to build unified models across diverse product categories. However, as the number of categories grows, its performance often degrades due to increasingly complex and heterogeneous normal distributions. To address this challenge, we propose DPDiff-AD, a \textbf{D}ual \textbf{P}rototype-conditioned \textbf{Diff}usion model for large-scale multi-class \textbf{A}nomaly \textbf{D}etection. DPDiff-AD models heterogeneous normal distributions through complementary local and global prototypes. Local prototypes capture representative fine-grained structural patterns via nearest-prototype aggregation, while global prototypes regulate holistic feature geometry through optimal transport regularization. Together, these dual-scale representations define a structured normality space. This space is refined through diffusion-based reconstruction conditioned on both local and global prototypes via prototype-aware attention. By jointly leveraging dual prototypes during generation, DPDiff-AD achieves precise normality modeling, preserves structured separability as category cardinality grows, and enables scalable anomaly discrimination. Extensive experiments across five benchmarks demonstrate the effectiveness and scalability of DPDiff-AD. On the 160-category large-scale dataset, it improves image- and pixel-level AUROC by 5.3 and 2.9 points over the previous state-of-the-art method Dinomaly+, while maintaining stable performance as category cardinality increases.
}


\maketitle

\section{Introduction}
Unsupervised anomaly detection (UAD) plays a pivotal role in industrial defect detection~\cite{bergmann2021mvtec}, medical image analysis~\cite{guo2023encoder}, and video surveillance~\cite{ramachandra2020survey}. It aims to identify and localize abnormal patterns using only normal training data \cite{cao2024survey}. While early research primarily focused on single-class settings\cite{bergmann2020uninformed,defard2021padim,li2021cutpaste}, modern deployment scenarios require unified models that operate across numerous heterogeneous categories within a shared framework\cite{you2022unified}. This transition from single-class to multi-class unsupervised anomaly detection (MUAD) fundamentally reshapes the learning problem: instead of modeling a single compact distribution, the model must represent increasingly diverse and multi-modal normal distributions within a shared latent space. As a result, scalability under category growth becomes a central challenge for practical large-scale MUAD deployment\cite{guo2025dinomaly,zhu2026real}.

Existing MUAD approaches can be broadly categorized into embedding-based, discrimination-based, and reconstruction-based paradigms. 
Embedding-based methods typically extract pretrained visual features and estimate anomaly scores through feature-space deviations or nearest-neighbor matching~\cite{roth2022towards,deng2022anomaly,gudovskiy2022cflow,lee2022cfa}. 
Discrimination-based methods learn decision boundaries by distinguishing normal samples from synthesized or augmented anomalies~\cite{zavrtanik2021draem,schluter2022natural,liu2023simplenet}. 
Reconstruction-based methods model normality by reconstructing input images or feature representations, where anomalies are expected to produce larger reconstruction discrepancies~\cite{you2022unified,deng2022anomaly,lu2023hierarchical,zhang2023destseg,gao2024learning,he2024diffusion,he2024mambaad,fan2025salvaging,guo2025dinomaly}. 
With the development of pretrained visual representations, transformer architectures~\cite{lu2023hierarchical,guo2025dinomaly,luo2025exploring}, and diffusion mechanisms~\cite{he2024diffusion,fuvcka2024transfusion,yao2024glad}, recent MUAD methods have achieved highly competitive and sometimes near-saturated performance on limited-category benchmarks, such as MVTec-AD~\cite{bergmann2021mvtec} with 15 categories and VisA~\cite{zou2022spot} with 12 categories.
In particular, reconstruction-based frameworks have shown remarkable effectiveness by enhancing global modeling and normality reconstruction in these settings. 
Collectively, these findings may create the impression that current MUAD approaches are already sufficiently mature, while leaving their scalability under category expansion insufficiently examined.

\begin{figure*}[t!]\vspace{0mm}
		\centering
		\includegraphics[width=90mm]{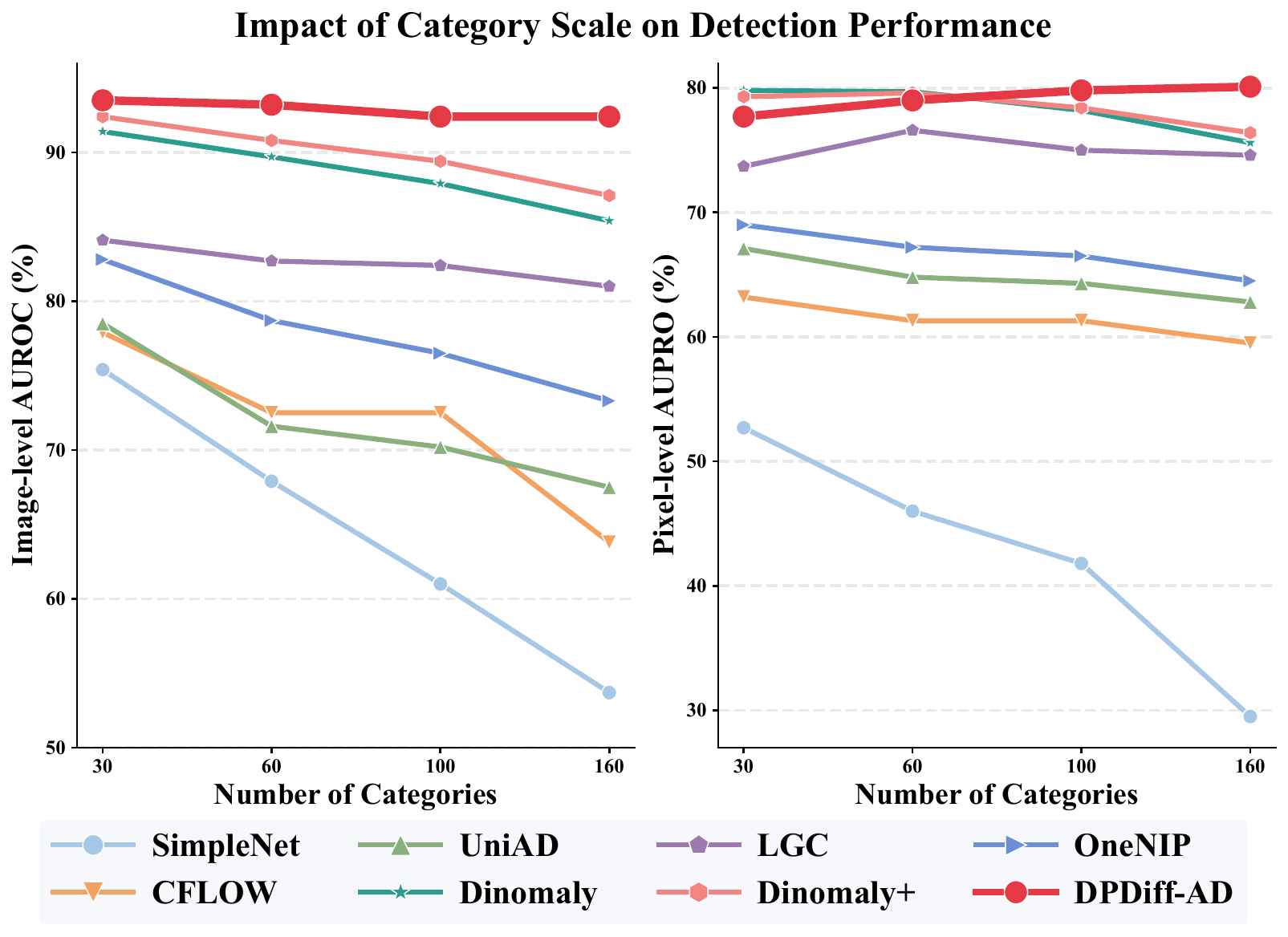} \vspace{0mm}
\caption{Impact of category scale on MUAD performance evaluated on Real-IAD Variety. Left: image-level AUROC (\%). Right: pixel-level AUPRO (\%). As the number of categories increases from 30 to 160, most existing MUAD models exhibit substantial performance degradation, while DPDiff-AD maintains consistently high performance.}
	\label{fig.trend}
	\end{figure*}

However, as the category scale expands from a few dozens to hundreds of heterogeneous classes, existing MUAD models exhibit substantial and systematic performance degradation\cite{zhu2026real}. As illustrated in Figure~\ref{fig.trend}, this phenomenon persists across methodological paradigms, including embedding-based approaches, discriminative frameworks, and reconstruction-based models—even those built upon powerful vision foundation backbones such as DINOv2\cite{oquab2023dinov2}. Despite enhanced representational capacity, performance consistently declines as category cardinality increases. These observations suggest that large-scale MUAD is not a straightforward extension of single-class UAD, but a structurally more demanding representation learning problem.

These phenomena arise from a deeper structural tension in large-scale MUAD. 
As the number of classes increases, unified models must organize increasingly heterogeneous normal patterns within a shared representation space, which may result in representational homogenization and blurred anomaly boundaries. 
This issue becomes particularly critical for reconstruction-based frameworks, where faithfully modeling complex multi-modal normal distributions across diverse categories requires sufficient reconstruction capacity. 
However, simply increasing reconstruction capacity is insufficient: excessive generalization may induce the ``identical shortcut''~\cite{gong2019memorizing,you2022unified,you2022adtr}, reconstructing anomalous patterns instead of projecting them toward normal manifolds, thereby diminishing residual-based detection signals. 
Together, these intertwined effects reveal a fundamental tension between structuring heterogeneous normal representations and accurately modeling complex normal distributions under large-scale category growth.

To address these challenges, we propose DPDiff-AD, a \textbf{D}ual \textbf{P}rototype-conditioned \textbf{Diff}usion model for large-scale multi-class \textbf{A}nomaly \textbf{D}etection. DPDiff-AD explicitly structures the normal representation space through complementary local and global constraints. At the local level, category-specific structural patterns are stabilized via nearest-prototype aggregation, which provides consistent anchors for fine-grained representation geometry. At the global level, feature distributions are  regularized through optimal transport alignment, enforcing cross-category consistency at the distribution scale. Together, these dual-scale mechanisms define a structured normality space that counteracts representation homogenization under category growth. Beyond representation stabilization, DPDiff-AD integrates diffusion-based generative modeling to faithfully capture complex multi-modal normal densities. By conditioning the diffusion process on both local and global prototypes via prototype-aware attention and injecting structured priors into latent alignment and diffusion dynamics, the framework guides reconstruction toward normal manifolds, mitigating ``identical shortcut'' and enhancing residual-based anomaly signals. Extensive experiments across five benchmarks demonstrate the effectiveness and scalability of DPDiff-AD. On the 160-category large-scale dataset, it improves image- and pixel-level AUROC by 5.3 and 2.9 points over the previous state-of-the-art method Dinomaly+. While existing unified models degrade under large-scale category growth, DPDiff-AD maintains stable performance, highlighting its superior scalability.

The main contributions of our work are summarized as follows:
\begin{itemize}
    \item \textbf{Structured Normality Modeling.}
    We introduce a dual-scale structured alignment mechanism that combines local aggregation and global optimal transport regularization to organize heterogeneous normal representations and maintain representation separability under increasing distributional diversity.

    \item \textbf{Dual Prototype-Conditioned Diffusion Model.}
Structured prototype priors are integrated into diffusion-based reconstruction by conditioning the denoising process on both local and global prototypes via prototype-aware attention, mitigating ``identical shortcut'' effects and enhancing anomaly discriminability in large-scale MUAD.

    \item \textbf{Large-Scale Evaluation.}
Comprehensive evaluations on five benchmarks, including the large-scale 160-category Real-IAD Variety dataset, demonstrate the effectiveness of DPDiff-AD. On Real-IAD Variety, DPDiff-AD improves image- and pixel-level AUROC by 5.3 and 2.9 points, respectively, over prior state-of-the-art approaches, while maintaining stable performance as category cardinality increases.

\end{itemize}

\section{Related Works}
\subsection{Multi-Class Unsupervised Anomaly Detection}

MUAD aims to detect anomalies across multiple object categories using a unified model. UniAD~\cite{you2022unified} first formalized this setting and demonstrated the feasibility of unified reconstruction-based modeling. Since then, a series of methods have been proposed to enhance MUAD, including diffusion-based approaches such as LafitE~\cite{yin2023lafite} and DiAD~\cite{he2024diffusion}, transformer-based frameworks such as ViTAD~\cite{zhang2023exploring} and HVQ-Trans~\cite{lu2023hierarchical}, state-space architectures such as MambaAD~\cite{he2024mambaad}, and foundation-model-based methods such as Dinomaly~\cite{guo2025dinomaly} and AACLIP~\cite{ma2025aa}. These methods achieve strong performance when the category scale is relatively limited, as commonly observed on benchmarks such as MVTec-AD and VisA. However, as evaluation shifts toward larger category spaces and richer intra-class variations, recent large-scale datasets such as Real-IAD Variety~\cite{zhu2026real} reveal that unified models still suffer from noticeable performance degradation under category expansion. This indicates that maintaining stable normal representations under large-scale category growth remains an open challenge for MUAD.

\subsection{Diffusion Models for Anomaly Detection}
Diffusion models~\cite{sohl2015deep,song2020score,ho2020denoising} have attracted increasing attention in anomaly detection due to their strong generative modeling and iterative denoising capabilities. 
Early diffusion-based anomaly detection methods mainly focused on restoring anomalous inputs toward anomaly-free appearances or improving anomaly localization through reconstruction discrepancies. 
AnoDDPM~\cite{wyatt2022anoddpm} introduces partial diffusion with simplex noise to restore anomalous medical images toward healthy appearances, while DiffAD~\cite{zhang2023unsupervised} employs latent diffusion with noisy condition embedding and interpolated channels to improve anomaly-free reconstruction and mitigate anomaly copying. 
DDAD~\cite{mousakhan2024anomaly} further develops target-conditioned denoising and combines pixel- and feature-level discrepancies for anomaly localization. 
More recent methods further explore diffusion-based anomaly detection in unified multi-class settings and introduce adaptive, localization-aware, or semantics-preserving denoising strategies. 
For example, TransFusion~\cite{fuvcka2024transfusion} jointly performs reconstruction and localization via transparency-based diffusion, GLAD~\cite{yao2024glad} adaptively adjusts denoising from both global and local perspectives, LafitE~\cite{yin2023lafite} conducts latent feature diffusion with feature editing to alleviate identity shortcuts, and DiAD~\cite{he2024diffusion} leverages stable diffusion priors with semantic guidance to preserve category semantics during anomaly reconstruction. 
However, existing diffusion-based methods mainly focus on improving reconstruction fidelity or localization quality, while the organization of heterogeneous normal representations in large-scale unified settings remains underexplored. 
To address this limitation, DPDiff-AD conditions diffusion reconstruction on dual-scale prototypes, thereby coupling diffusion-based normality reconstruction with prototype-based representation stabilization under category growth.

\subsection{Prototype Learning for Anomaly Detection}

Prototype learning~\cite{snell2017prototypical} represents data patterns as anchors in a metric space and has been widely adopted in metric and few-shot learning. 
In anomaly detection, PatchCore~\cite{roth2022towards} builds memory banks of normal patch features for nearest-neighbor scoring, while FastRef~\cite{tian2025fastref} enhances prototype-based detection through test-time refinement. 
HVQ-Transformer~\cite{lu2023hierarchical} incorporates vector-quantized prototypes into a transformer-based reconstruction framework, and INP-Former~\cite{luo2025exploring} dynamically extracts intrinsic normal prototypes from the test image to improve alignment in unified settings. 
Despite these advances, existing prototype-based methods mainly rely on local memory retrieval, vector quantization, or image-specific prototype extraction, while the joint organization of local structural patterns and global distribution geometry under large-scale category diversity remains insufficiently explored. 
As category diversity increases, this limitation makes it difficult to maintain coherent normal representations and stable anomaly boundaries within a shared representation space. 
In contrast, DPDiff-AD introduces complementary local and global prototypes to structure heterogeneous normal representations and further uses them to guide diffusion-based reconstruction.

\section{Preliminaries}
\subsection{Diffusion Models}
Diffusion probabilistic models~\cite{sohl2015deep,song2020score,ho2020denoising}, particularly the widely adopted Denoising Diffusion Probabilistic Model (DDPM)~\cite{ho2020denoising}, consist of two key components: a forward diffusion process, in which the input \(\boldsymbol{x}_0\) is gradually corrupted by Gaussian noise over \(T\) timesteps, and a reverse denoising process, which learns to reconstruct the original noise-free data from the noisy observations. 

Formally, the forward process can be expressed as:
    \begin{equation} \label{originalDDPM1}
        \begin{array}{c}
            q(\boldsymbol{x}_{1:T} \mid \boldsymbol{x}_{0}) := \prod_{t=1}^T q(\boldsymbol{x}_{t} \mid \boldsymbol{x}_{t-1}), \\[2mm]  
            q(\boldsymbol{x}_{t} \mid \boldsymbol{x}_{t-1}):= \mathcal{N}(\sqrt{\alpha_t} \boldsymbol{x}_{t-1}, (1-\alpha_t) \boldsymbol{I})
        \end{array}
    \end{equation}
    
where \(\alpha_t := 1-\beta_t\), and \(\beta_t \in (0,1)\) controls the noise level. In practice, \(\boldsymbol{x}_t\) can also be sampled directly from \(\boldsymbol{x}_0\) via:
    \begin{equation} 
    q(\boldsymbol{x}_{t} \mid \boldsymbol{x}_{0}) = \mathcal{N}(\sqrt{\bar{\alpha}_t} \boldsymbol{x}_{0}, (1-\bar{\alpha}_t) \boldsymbol{I})
    \end{equation}
where \(\bar{\alpha}_t := \prod_{s=1}^{t} \alpha_s\). 
The reverse process reconstructs \(\boldsymbol{x}_0\) from \(\boldsymbol{x}_t\) through a denoising procedure:
    \begin{equation} \label{originalDDPM2}
        \begin{array}{c}
            p_\theta(\boldsymbol{x}_{0:T}) := p(\boldsymbol{x}_T) \prod_{t=1}^T p_\theta(\boldsymbol{x}_{t-1} \mid \boldsymbol{x}_{t}), \\[2mm] 
            p_\theta(\boldsymbol{x}_{t-1} \mid \boldsymbol{x}_{t}) := \mathcal{N}(\boldsymbol{\mu}_\theta(\boldsymbol{x}_{t},t), \tilde{\beta}_t\boldsymbol{I})
        \end{array}
    \end{equation}
In DDPM, the conditional distribution \(p_\theta(\boldsymbol{x}_{t-1} \mid \boldsymbol{x}_t)\) is parameterized as:
    \begin{equation} \label{originalDDPM3}
        \begin{array}{c}
    \boldsymbol{\mu}_\theta(\boldsymbol{x}_{t},t)=\frac{1}{\sqrt{\alpha_t}}\left(\boldsymbol{x}_{t}-\frac{\beta_t}{\sqrt{1-\bar{\alpha}_t}}\boldsymbol{\epsilon}_\theta(\boldsymbol{x}_{t},t)\right),  
            \tilde{\beta}_t=  \frac{1-\bar{\alpha}_{t-1}}{1-\bar{\alpha}_t}\beta_t 
        \end{array}
    \end{equation}
where \(\boldsymbol{\epsilon}_\theta\) denotes the learnable denoising function, typically implemented using a UNet~\cite{ronneberger2015u,dhariwal2021diffusion} or DiT~\cite{peebles2023scalable} backbone, and trained by minimizing the following objective:
    \begin{equation} \label{originalDDPM4}
            \mathop{\min}\limits_{\theta} \mathcal{L}(\theta) := \mathbb{E}_{\boldsymbol{x}_0,t,\boldsymbol{\epsilon}} [\gamma(t) \ \Vert \boldsymbol{\epsilon} - \boldsymbol{\epsilon}_\theta(\boldsymbol{x}_{t},t) \Vert_2^2]
    \end{equation}
where \(\gamma(t)\) is a timestep-dependent weight that balances contributions from different noise levels during training.

\subsection{Optimal Transport}
Optimal transport (OT) quantifies the discrepancy between probability distributions by solving a minimal-cost transport problem~\cite{peyre2017computational}. 
Consider two discrete distributions
\(
P = \sum_{i=1}^{m} a_i \delta_{x_i} \quad \text{and} \quad
Q = \sum_{j=1}^{n} b_j \delta_{y_j},
\)
where \(a \in \Delta^m\) and \(b \in \Delta^n\) lie in the probability simplex, and \(\delta_x\) denotes the Dirac function at point \(x\). 
The OT distance between \(P\) and \(Q\) is defined as

\begin{equation}
\mathrm{OT}(P, Q) = \min_{T \in \Pi(a,b)} \langle T, C \rangle,
\end{equation}

where \(C_{ij} = c(x_i, y_j)\) is the transport cost, and 

\[
\Pi(a,b) = \{T \ge 0 \mid T\mathbf{1} = a,\; T^\top \mathbf{1} = b\}
\]

denotes the set of admissible transport plans. 

To reduce computational cost, we employ the entropy-regularized OT formulation:

\begin{equation}
\mathrm{OT}_\varepsilon(P, Q) = \min_{T \in \Pi(a,b)} \langle T, C \rangle + \varepsilon \sum_{i,j} T_{ij} \log T_{ij},
\end{equation}

which can be efficiently solved using the Sinkhorn algorithm~\cite{peyre2017computational}.

\section{Methodology}
\begin{figure*}[t!]\vspace{0mm}
		\centering
		\includegraphics[width=130mm]{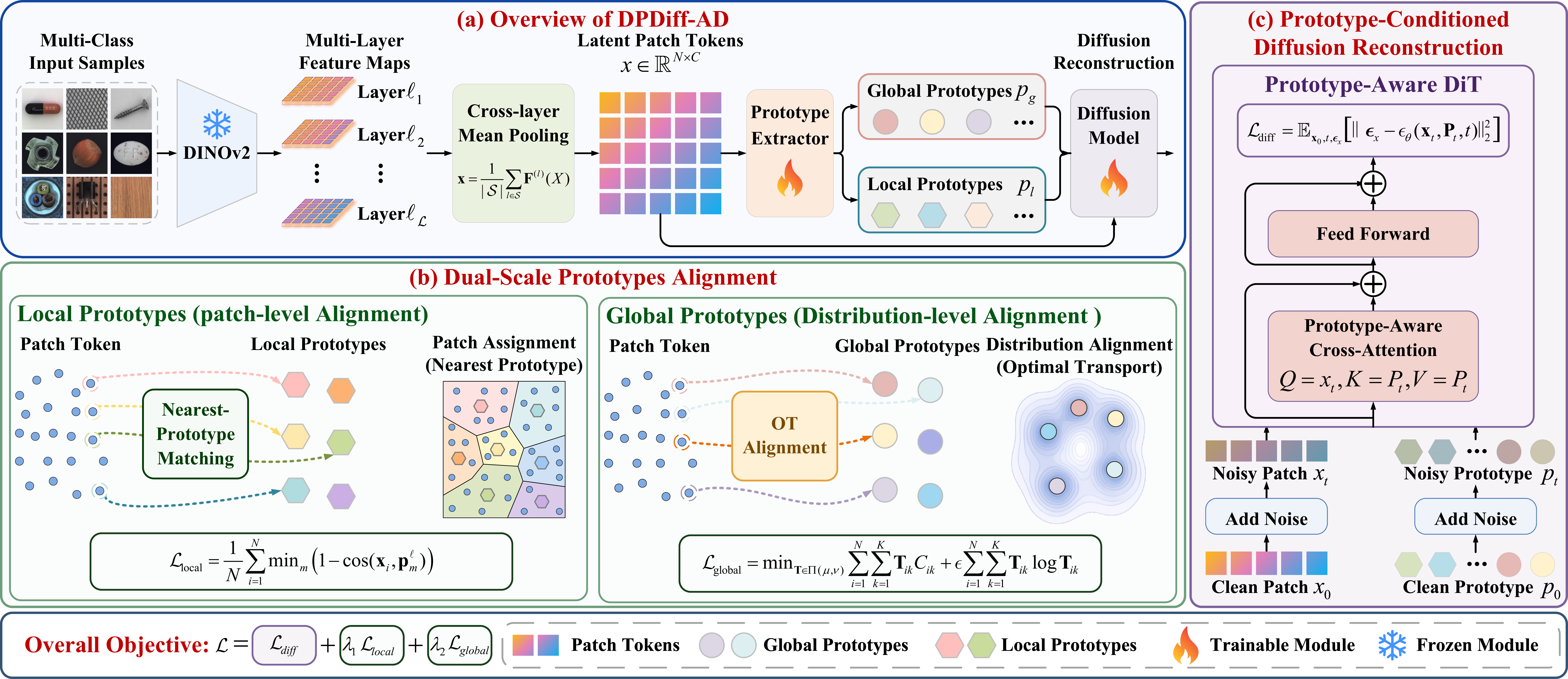} \vspace{0mm}
\caption{
Overview of DPDiff-AD for large-scale MUAD. 
(a) The overall pipeline extracts multi-layer DINOv2 features from multi-class normal samples, aggregates them into latent patch tokens, and learns dual-scale prototypes to structure the latent space. 
The latent patch tokens and dual-scale prototypes are then jointly fed into the diffusion model for feature reconstruction. 
(b) Dual-scale prototype alignment constrains feature representations from both patch-level and distribution-level perspectives, stabilizing the latent space under large category diversity. 
(c) Prototype-conditioned diffusion reconstruction employs a prototype-aware DiT with cross-attention and feed-forward layers to reconstruct features under prototype guidance. 
The overall objective jointly optimizes diffusion denoising, local prototype alignment, and global distribution alignment for robust anomaly detection across diverse categories.
}
	\label{fig.fig}
	\end{figure*}

We propose DPDiff-AD, a dual prototype-conditioned diffusion model for large-scale MUAD. 
As illustrated in Fig.~\ref{fig.fig}, DPDiff-AD first extracts hierarchical feature representations using a pretrained DINOv2 backbone, and then introduces dual-scale prototypes to regularize the latent space from both local and global perspectives. 
Specifically, local prototypes capture representative patch-level normal patterns, while global prototypes align the overall feature distribution via optimal transport. 
These structured prototypes are further incorporated into the diffusion reconstruction process through prototype-aware attention, enabling the denoising network to selectively exploit prototype-guided normal priors and guide anomalous features toward the learned normal manifold. 
By coupling dual-scale prototype alignment with diffusion-based normality modeling, DPDiff-AD enables stable anomaly discrimination under extensive category diversity.

\subsection{Representation Stabilization via Dual-Scale Prototypes}

Large-scale MUAD requires modeling normal samples from numerous heterogeneous categories within a shared feature space. 
However, such samples often exhibit highly complex and multi-modal distributions, leading to fragmented representations and entangled decision boundaries in the latent space. 
To address this challenge, we introduce a dual-scale prototype mechanism that regularizes latent features from both local and global perspectives. 
The local prototypes capture representative patch-level normal patterns, while the global prototypes provide distribution-level anchors to stabilize the overall feature geometry.

\paragraph{Hierarchical Feature Construction.}
Given an input image $X \in \mathbb{R}^{H \times W \times 3}$, 
we employ a pretrained DINOv2 backbone to extract latent feature representations from multiple transformer layers:
\begin{equation}
\mathbf{F}(X) =
\{\mathbf{F}^{(1)}(X), \mathbf{F}^{(2)}(X), \dots, \mathbf{F}^{(L)}(X)\},
\end{equation}
where $\mathbf{F}^{(l)}(X) \in \mathbb{R}^{N \times C}$ denotes the latent feature representation extracted from the $l$-th transformer block.

Features from different transformer layers capture complementary semantic information. 
To integrate these multi-layer representations, we select a subset of transformer layers $\mathcal{S}$ and aggregate their outputs via cross-layer mean pooling:
\begin{equation}
\mathbf{x} =
\frac{1}{|\mathcal{S}|}
\sum_{l \in \mathcal{S}}
\mathbf{F}^{(l)}(X),
\end{equation}
where $\mathbf{x} \in \mathbb{R}^{N \times C}$ denotes the aggregated latent feature representation.

\paragraph{Local Prototype Aggregation.}
To capture representative patch-level normal patterns and stabilize local feature structures in the latent space, we introduce a set of $M$ learnable local prototypes $\mathbf{P}_{\ell} = \{\mathbf{p}_{m}^{\ell}\}_{m=1}^{M}$, where $\mathbf{p}_{m}^{\ell} \in \mathbb{R}^{C}$.

Instead of directly clustering token features, we learn local prototypes through a Transformer layer that aggregates token representations via cross-attention. 
Given latent tokens $\mathbf{x} = \{\mathbf{x}_{i}\}_{i=1}^{N}$ and learnable prototype queries $\mathbf{T}$, the local prototypes are updated as
\begin{equation}
\mathbf{T}' =
\mathrm{CrossAttn}(\mathbf{T}, \mathbf{x}, \mathbf{x}) + \mathbf{T},
\quad
\mathbf{P}_{\ell} =
\mathrm{FFN}(\mathbf{T}') + \mathbf{T}'.
\end{equation}

To enforce local consistency between tokens and prototypes, we introduce a cosine-based regularization that associates each token with its nearest local prototype:
\begin{equation}
\mathcal{L}_{\mathrm{local}} =
\frac{1}{N}
\sum_{i=1}^{N}
\min_{m}
\left(1 - \cos(\mathbf{x}_{i}, \mathbf{p}_{m}^{\ell})\right).
\end{equation}

This objective encourages local prototypes to capture dominant patch-level normal patterns while serving as a soft geometric constraint, rather than enforcing hard cluster assignments.

\paragraph{Global Distribution Alignment via Optimal Transport.}

While local prototypes capture representative patch-level structures, their influence is mainly limited to constraining feature geometry at a local scale. 
In large-scale multi-class settings, normal samples from diverse categories often result in fragmented and heterogeneous feature distributions in the latent space. 
To regularize the global structure of this space, we introduce a set of global prototypes $\mathbf{P}_{g}=\{\mathbf{p}_{k}^{g}\}_{k=1}^{K}$ that serve as distribution-level anchors for normal patterns.

These prototypes are extracted using a Transformer module with the same architecture as the local prototype extractor but with independent parameters. 
Unlike local prototypes, which are optimized via token-level alignment, global prototypes aim to capture the overall geometry of the feature distribution and provide global structural guidance for latent representations.

Formally, we treat the token features $\mathbf{x}=\{\mathbf{x}_{i}\}_{i=1}^{N}$ and the global prototypes $\mathbf{P}_{g}$ as empirical distributions:
\begin{equation}
P_{\mathbf{x}} =
\frac{1}{N} \sum_{i=1}^{N} \delta_{\mathbf{x}_{i}},
\quad
P_{\mathbf{g}} =
\frac{1}{K} \sum_{k=1}^{K} \delta_{\mathbf{p}_{k}^{g}},
\end{equation}
where $\delta$ denotes the Dirac measure. 
To align these two distributions, we employ entropy-regularized optimal transport:
\begin{equation}
\mathcal{L}_{\mathrm{global}}
=
\min_{\mathbf{T}\in\Pi(\mu,\nu)}
\sum_{i=1}^{N}\sum_{k=1}^{K} \mathbf{T}_{ik} C_{ik}
+
\epsilon
\sum_{i=1}^{N}\sum_{k=1}^{K} \mathbf{T}_{ik}\log \mathbf{T}_{ik},
\end{equation}
where \(\mathbf{T}\) denotes the transport plan, \(\Pi(\mu,\nu)\) is the set of admissible couplings between the token distribution and the global prototype distribution, and \(C_{ik}=\|\mathbf{x}_{i}-\mathbf{p}_{k}^{g}\|_2^2\) defines the transport cost matrix.

This objective aligns token features with global prototypes at the distribution level, thereby regularizing the global geometry of the latent space and mitigating representation fragmentation under large-scale category diversity.

\subsection{Prototype-Conditioned Diffusion Reconstruction}
The learned dual-scale prototypes provide structured anchors that characterize representative normal patterns in the latent space. 
To model the complex and multi-modal normal distributions arising in large-scale MUAD, we incorporate these prototypes into a diffusion reconstruction framework. 
Specifically, the latent patch tokens and dual-scale prototypes are jointly perturbed and then reconstructed by a prototype-aware diffusion transformer, where prototype-aware attention further refines the interaction between tokens and prototypes. 
This enables the model to recover features toward a structured normal manifold consistent with the prototype-induced latent geometry.

\paragraph{Forward Diffusion Process}
Let $\mathbf{x}_0 \in \mathbb{R}^{N \times C}$ denote the latent feature tokens extracted from the backbone, and let
$\mathbf{P}_0 = [\mathbf{P}_{\ell}; \mathbf{P}_{g}] \in \mathbb{R}^{(M+K) \times C}$
denote the concatenated dual-scale prototypes.

To maintain consistent noise levels between latent tokens and their prototype-guided structural priors, Gaussian noise is progressively injected into both feature tokens and prototypes at timestep $t$:
\begin{equation}
\mathbf{x}_t =
\sqrt{\bar{\alpha}_t}\mathbf{x}_0 +
\sqrt{1-\bar{\alpha}_t}\boldsymbol{\epsilon}_x,
\quad
\boldsymbol{\epsilon}_x \sim \mathcal{N}(0,I),
\end{equation}

\begin{equation}
\mathbf{P}_t =
\sqrt{\bar{\alpha}_t}\mathbf{P}_0 +
\sqrt{1-\bar{\alpha}_t}\boldsymbol{\epsilon}_p,
\quad
\boldsymbol{\epsilon}_p \sim \mathcal{N}(0,I),
\end{equation}
where $\{\bar{\alpha}_t\}$ denotes the predefined noise schedule.

\paragraph{Prototype-Aware Diffusion Transformer}
Given the noisy feature tokens $\mathbf{x}_t$ and noisy prototypes $\mathbf{P}_t$, we implement the denoising network with a Diffusion Transformer (DiT). 
To explicitly leverage the prototype-induced latent structure during denoising, we incorporate prototype-aware attention into each DiT block, enabling feature tokens to interact with prototype representations.

Specifically, the prototype-aware interaction is formulated as:
\begin{equation}
\mathbf{x}_t' =
\mathbf{x}_t +
\mathrm{CrossAttn}(\mathbf{x}_t,\mathbf{P}_t,\mathbf{P}_t),
\end{equation}
where
\begin{equation}
\mathrm{CrossAttn}(\mathbf{Q},\mathbf{K},\mathbf{V})
=
\mathrm{Softmax}
\left(
\frac{\mathbf{Q}\mathbf{K}^\top}{\sqrt{d}}
\right)\mathbf{V}.
\end{equation}

Through this mechanism, feature tokens dynamically attend to prototype representations, allowing the denoising process to be guided by the structured geometry of the latent space. 
This prototype-aware interaction encourages reconstructed features to remain consistent with dominant normal patterns captured by the prototypes. 
Time-dependent conditioning is incorporated through adaptive layer normalization (adaLN), which modulates intermediate activations based on timestep embeddings.

\paragraph{Training Objective.}
Following the standard DDPM formulation~\cite{ho2020denoising},
the denoising network $\epsilon_\theta(\cdot)$ is trained to
predict the injected Gaussian noise of feature tokens, conditioned on
the noisy prototypes and timestep $t$.
The diffusion model is optimized using the standard
noise-prediction objective:
\begin{equation}
\mathcal{L}_{\mathrm{diff}}
=
\mathbb{E}_{\mathbf{x}_0,t,\boldsymbol{\epsilon}_{x}}
\Big[
\|
\boldsymbol{\epsilon}_{x}
-
\epsilon_\theta(\mathbf{x}_t,\mathbf{P}_t,t)
\|_2^2
\Big].
\end{equation}

\paragraph{Overall Objective.}
The overall training objective integrates three complementary
components: diffusion denoising, local geometric
regularization, and global distribution alignment:
\begin{equation}
\mathcal{L}
=
\mathcal{L}_{\mathrm{diff}}
+
\lambda_1 \mathcal{L}_{\mathrm{local}}
+
\lambda_2 \mathcal{L}_{\mathrm{global}},
\end{equation}
where $\lambda_1$ and $\lambda_2$ balance the contributions
of local prototype regularization and global distribution
alignment.

\subsection{Anomaly Localization and Detection}

During inference, the latent feature tokens \(\mathbf{x}_0\) and prototypes \(\mathbf{P}_0\) are first perturbed with Gaussian noise at a fixed timestep \(t\), yielding \(\mathbf{x}_t\) and \(\mathbf{P}_t\). 
Starting from \(\mathbf{x}_t\), the diffusion transformer performs reverse denoising conditioned on the noisy prototypes. 
At each reverse timestep \(s=t,\dots,1\), the denoising network predicts the noise component as
\begin{equation}
\hat{\boldsymbol{\epsilon}}_s
=
\boldsymbol{\epsilon}_{\theta}(\mathbf{x}_s,\mathbf{P}_s,s),
\end{equation}
where \(\mathbf{P}_s\) denotes the prototype condition at timestep \(s\), obtained by perturbing \(\mathbf{P}_0\) according to the same noise schedule. 
The previous latent state is then estimated by
\begin{equation}
\mathbf{x}_{s-1}
=
\frac{1}{\sqrt{\alpha_s}}
\left(
\mathbf{x}_s
-
\frac{\beta_s}{\sqrt{1-\bar{\alpha}_s}}
\hat{\boldsymbol{\epsilon}}_s
\right)
+
\sigma_s \mathbf{z},
\quad
\mathbf{z}\sim\mathcal{N}(\mathbf{0},\mathbf{I}),
\end{equation}
where \(\sigma_s^2=\tilde{\beta}_s\) for DDPM sampling, \(\sigma_s\) controls the stochasticity of the reverse process, and no additional noise is injected at the final denoising step. 
After iterative denoising, the reconstructed feature tokens are obtained as \(\hat{\mathbf{x}}_0 \in \mathbb{R}^{N \times C}\). 
In practice, fast sampling strategies such as DDIM~\cite{song2020denoising} can be adopted to reduce the number of denoising steps while maintaining reconstruction quality.

The anomaly score of each token is computed by the channel-wise squared \(\ell_2\) reconstruction discrepancy between the reconstructed and original features:
\begin{equation}
s_i =
\left\|
\hat{\mathbf{x}}_{0,i} - \mathbf{x}_{0,i}
\right\|_2^2,
\quad i=1,\dots,N.
\end{equation}
The token-level anomaly scores are then reshaped into a spatial anomaly map:
\begin{equation}
S =
\mathrm{Reshape}(\{s_i\}_{i=1}^{N})
\in \mathbb{R}^{h \times w},
\quad N=hw.
\end{equation}

Since reconstruction is performed in the backbone feature space, the anomaly map \(S\) is upsampled to the input image resolution via bilinear interpolation, yielding \(S_{\uparrow}\). 
For image-level anomaly detection, the upsampled anomaly map is aggregated using average pooling followed by a max operator:
\begin{equation}
s_{\mathrm{img}}
=
\max\left(\mathrm{AvgPool}(S_{\uparrow})\right).
\end{equation}

\section{Experiments and Analysis}

\subsection{Experimental Settings}
\subsubsection{Datasets}
We evaluate our method on five industrial anomaly detection benchmarks, including the large-scale Real-IAD Variety benchmark and four widely used benchmarks.

\textbf{Real-IAD Variety}~\cite{zhu2026real} is a recently proposed large-scale industrial anomaly detection benchmark containing 198,950 high-resolution images from 160 object categories. It covers 28 industries, 24 material types, 22 color variations, and 23 defect types, providing substantial category diversity and intra-class variation for evaluating the scalability of MUAD methods. Following its official category-scale protocol, we evaluate scalability on S1, S2, S3, and the full setting, which contain 30, 60, 100, and 160 categories, respectively. The S1/S2/S3 partitions are defined by the original Real-IAD authors using randomized category selection to reduce potential color and material biases, and the corresponding category breakdown is provided in Appendix~\ref{append_S1S2S3_category}.

In addition to Real-IAD Variety, we evaluate our method on four widely used industrial anomaly detection benchmarks to facilitate fair comparisons with prior work and demonstrate its general applicability. \textbf{MVTec-AD}~\cite{bergmann2021mvtec} contains 15 object and texture categories with high-resolution images for unsupervised anomaly detection. \textbf{VisA}~\cite{zou2022spot} consists of 12 object categories with 9,621 normal images and 1,200 anomalous images. \textbf{MPDD}~\cite{jezek2021deep} contains six categories of painted metal parts, including 888 normal training images and 458 test images. \textbf{Real-IAD}~\cite{wang2024real} contains approximately 150,000 images across 30 industrial categories.

\subsubsection{Evaluation metrics}
Following prior works~\cite{zhang2023exploring,he2024mambaad}, we evaluate anomaly detection and localization performance using both image-level and pixel-level metrics. For image-level anomaly detection, we report the Area Under the Receiver Operating Characteristic Curve (AUROC), Average Precision (AP)~\cite{zavrtanik2021draem}, and F$_1$-max~\cite{zou2022spot}. For pixel-level anomaly localization, we report AUROC, AP, F$_1$-max, and the Area Under the Per-Region-Overlap (AUPRO)~\cite{bergmann2020uninformed}. To provide an overall assessment of model performance, we further compute the mean of these seven metrics, referred to as mAD~\cite{zhang2023exploring}. For efficiency evaluation, we report the inference speed in frames per second (FPS).

\subsubsection{Implementation Details}
We use a pretrained DINOv2-B backbone~\cite{darcet2023vision} and extract intermediate features from transformer layers 6--11. For the prototype module, we use $M=8$ local prototypes and $K=2$ global prototypes to capture representative patch-level structures and global distribution geometry, respectively. The diffusion process uses $T=1000$ timesteps with a linear noise schedule ($\beta_1=10^{-4}$, $\beta_T=0.02$), following the standard DDPM setting~\cite{ho2020denoising}. 
All images are resized to $448\times448$. 
The hyperparameters $\lambda_1$ and $\lambda_2$ are set to 0.2 and 0.001, respectively. The model is trained using the Adam optimizer~\cite{kingma2014adam} with a learning rate of $1\times10^{-4}$ and a batch size of 32. All experiments are conducted on an NVIDIA RTX 4090 GPU with 24GB of VRAM.

\subsection{Evaluations on the Large-Scale Industrial Benchmark: Real-IAD Variety}

We begin by evaluating  DPDiff-AD on the large-scale industrial benchmark Real-IAD Variety, which comprises 160 object categories and introduces substantial challenges due to extreme category diversity and pronounced intra-class variation. This benchmark provides a realistic testbed for assessing the scalability and robustness of MUAD methods in complex industrial scenarios.

\begin{table*}[t]
\centering
\caption{Quantitative MUAD performance comparison on the large-scale Real-IAD Variety dataset containing 160 categories. Dis., Emb., and Rec. denote discrimination-based, embedding-based, and reconstruction-based methods,
respectively. The best results are indicated in \textbf{bold}.}
\label{Real-IAD-Variety}

\resizebox{0.95\textwidth}{!}{
\begin{tabular}{llccc cccc c}
\toprule

\multirow{2}{*}{\textbf{Paradigms}} &
\multirow{2}{*}{\textbf{Methods}} &
\multicolumn{3}{c}{\textbf{Image-level}} &
\multicolumn{4}{c}{\textbf{Pixel-level}} &
\multirow{2}{*}{\textbf{mAD}} \\

\cmidrule(lr){3-5} \cmidrule(lr){6-9}

 & &
\textbf{AUROC} & \textbf{AP} & \textbf{F$_1$-max} &
\textbf{AUROC} & \textbf{AP} & \textbf{F$_1$-max} & \textbf{AUPRO} & \\

\midrule

\multirow{2}{*}{Dis.}  
& DRAEM~\cite{zavrtanik2021draem} & 54.9 & 89.3 & 93.1 & 53.7 & 9.4 & 8.8 & 21.6 & 47.3 \\
& SimpleNet~\cite{liu2023simplenet} & 53.7 & 88.5 & 93.0 & 62.6 & 4.8 & 8.8 & 29.5 & 48.7 \\

\midrule

\multirow{2}{*}{Emb.}  
& CFA~\cite{lee2022cfa} & 52.9 & 88.5 & 93.0 & 52.9 & 2.7 & 5.6 & 14.1 & 44.2 \\
& CFLOW~\cite{gudovskiy2022cflow} & 63.8 & 91.7 & 93.1 & 85.2 & 14.8 & 20.0 & 59.5 & 61.2 \\

\midrule

\multirow{9}{*}{Rec.} 
& UniAD~\cite{you2022unified} & 67.5 & 92.4 & 93.3 & 87.1 & 18.0 & 24.0 & 62.8 & 63.6 \\
& OneNIP~\cite{gao2024learning} & 73.3 & 94.2 & 93.5 & 88.3 & 24.0 & 29.4 & 64.5 & 66.7 \\
& RD4AD~\cite{deng2022anomaly} & 72.8 & 93.9 & 93.6 & 88.9 & 22.3 & 28.1 & 68.9 & 66.9 \\
& DeSTSeg~\cite{zhang2023destseg} & 75.0 & 95.0 & 93.2 & 65.6 & 37.2 & 31.2 & 37.7 & 62.1 \\

& MambaAD~\cite{he2024mambaad} & 81.7 & 96.2 & 93.9 & 91.2 & 33.1 & 37.7 & 73.5 & 72.5 \\

& ViTAD~\cite{zhang2023exploring} & 79.4 & 95.6 & 93.8 & 91.0 & 31.9 & 36.5 & 70.7 & 71.3 \\

& LGC~\cite{fan2025salvaging} & 81.0 & 96.0 & 94.0 & 91.7 & 31.9 & 37.0 & 74.6 & 72.3 \\

& HVQ-Trans~\cite{lu2023hierarchical} & 84.7 & 97.3 & 94.7 & 89.4 & 39.4 & 42.5 & 73.2 & 74.5 \\

& Dinomaly~\cite{guo2025dinomaly} & 85.4 & 97.2 & 94.5 & 91.5 & 42.8 & 45.8 & 75.6 & 76.1 \\
& Dinomaly+~\cite{guo2025dinomaly} & 87.1 & 97.3 & 94.0 & 91.9 & 49.7 & 49.2 & 76.4 & 77.9 \\
\rowcolor[gray]{0.95}
& \textbf{DPDiff-AD} & \textbf{92.4} & \textbf{98.5} & \textbf{95.8} & \textbf{94.8} & \textbf{50.1} & \textbf{50.7} & \textbf{80.1} & \textbf{80.4} \\

\bottomrule
\end{tabular}
}

\end{table*}
\subsubsection{Overall Performance on the Real-IAD Variety Benchmark}

As shown in Table~\ref{Real-IAD-Variety}, DPDiff-AD achieves the best overall performance on the large-scale Real-IAD Variety dataset containing 160 object categories. Compared with existing discrimination-based, embedding-based, and reconstruction-based approaches, DPDiff-AD achieves the best results across all reported metrics, demonstrating its effectiveness in large-scale MUAD.

Large-scale category diversity in realistic industrial scenarios introduces significant challenges, as the distribution of normal samples becomes highly heterogeneous and difficult to characterize. Under this challenging setting, DPDiff-AD attains an image-level AUROC of 92.4\% and a pixel-level AUROC of 94.8\%, substantially surpassing the strongest competing method, Dinomaly+, by 5.3 and 2.9 points, respectively. Furthermore, DPDiff-AD achieves consistent improvements across all evaluation criteria, including F$_1$-max and AUPRO. Notably, when aggregating all seven image-level and pixel-level evaluation metrics, DPDiff-AD yields an average performance gain of approximately 2.5 points over Dinomaly+, highlighting its balanced advantage in anomaly detection accuracy and localization quality.

These results indicate that the proposed DPDiff-AD framework effectively models complex and heterogeneous normal distributions under large-scale category diversity, thereby enabling robust and scalable anomaly detection across diverse industrial objects. Detailed per-category results on Real-IAD Variety are provided in Appendix~\ref{append Each Category}.

\begin{table*}[t]
\centering
\caption{Scalability evaluation of MUAD methods under increasing category scales on Real-IAD Variety. 
$\Delta$ denotes the performance range, computed as the absolute difference between the maximum and minimum values across the four category-scale settings, i.e., S1, S2, S3, and Full. A smaller $\Delta$ indicates lower sensitivity to category expansion and better scalability.}
\label{tab:Scalability evaluation}

\setlength{\tabcolsep}{3.5pt}

\resizebox{0.95\columnwidth}{!}{
\begin{tabular}{lccccc cccc c}
\toprule

\multirow{2}{*}{\textbf{Methods}} &
\multirow{2}{*}{\textbf{Datasets}} &
\multirow{2}{*}{\textbf{Classes}} &
\multicolumn{3}{c}{\textbf{Image-level}} &
\multicolumn{4}{c}{\textbf{Pixel-level}} &
\multirow{2}{*}{\textbf{mAD}} \\

\cmidrule(lr){4-6}
\cmidrule(lr){7-10}

& & &
\textbf{AUROC} &
\textbf{AP} &
\textbf{F$_1$-max} &
\textbf{AUROC} &
\textbf{AP} &
\textbf{F$_1$-max} &
\textbf{AUPRO} &
\\

\midrule

\multirow{5}{*}{SimpleNet~\cite{liu2023simplenet}}
& Real-IAD Variety S1 & 30 & 75.4 & 95.2 & 93.7 & 83.4 & 22.1 & 26.6 & 52.7 & 64.2 \\
& Real-IAD Variety S2 & 60 & 67.9 & 93.2 & 93.2 & 80.4 & 14.4 & 19.2 & 46.0 & 59.2 \\
& Real-IAD Variety S3 & 100 & 61.0 & 91.0 & 93.1 & 78.0 & 9.2 & 13.8 & 41.8 & 55.4 \\
& Real-IAD Variety & 160 & 53.7 & 88.5 & 93.0 & 62.6 & 4.8 & 8.8 & 29.5 & 48.7 \\
\rowcolor[gray]{0.95}
\multicolumn{3}{c}{$\Delta \downarrow$} & 21.7 & 6.7 & \textbf{0.7} & 20.8 & 17.3 & 17.8 & 23.2 & 15.5 \\

\midrule

\multirow{5}{*}{CFLOW~\cite{gudovskiy2022cflow}}
& Real-IAD Variety S1 & 30 & 77.9 & 95.4 & 93.8 & 87.5 & 27.6 & 26.6 & 63.2 & 67.4 \\
& Real-IAD Variety S2 & 60 & 72.5 & 94.1 & 93.3 & 86.7 & 23.6 & 24.5 & 61.3 & 65.1 \\
& Real-IAD Variety S3 & 100 & 72.5 & 94.1 & 93.3 & 86.7 & 23.6 & 24.5 & 61.3 & 65.1 \\
& Real-IAD Variety & 160 & 63.8 & 91.7 & 93.1 & 85.2 & 14.8 & 20.0 & 59.5 & 61.2 \\
\rowcolor[gray]{0.95}
\multicolumn{3}{c}{$\Delta \downarrow$} & 14.1 & 3.7 & \textbf{0.7} & 2.3 & 12.8 & 6.6 & 3.7 & 6.3 \\

\midrule

\multirow{5}{*}{UniAD~\cite{you2022unified}}
& Real-IAD Variety S1 & 30 & 78.5 & 95.4 & 94.2 & 88.9 & 29.5 & 34.1 & 67.1 & 69.7 \\
& Real-IAD Variety S2 & 60 & 71.6 & 93.5 & 93.7 & 87.3 & 21.7 & 27.6 & 64.8 & 65.7 \\
& Real-IAD Variety S3 & 100 & 70.2 & 93.1 & 93.5 & 87.6 & 19.0 & 25.0 & 64.3 & 64.7 \\
& Real-IAD Variety & 160 & 67.5 & 92.4 & 93.3 & 87.1 & 18.0 & 24.0 & 62.8 & 63.6 \\
\rowcolor[gray]{0.95}
\multicolumn{3}{c}{$\Delta \downarrow$} & 11.0 & 3.0 & 0.9 & 1.8 & 11.5 & 10.1 & 4.3 & 6.1 \\

\midrule

\multirow{5}{*}{OneNIP~\cite{gao2024learning}}
& Real-IAD Variety S1 & 30 & 82.8 & 96.7 & 94.4 & 90.1 & 36.7 & 39.1 & 69.0 & 72.7 \\
& Real-IAD Variety S2 & 60 & 78.7 & 95.6 & 94.0 & 89.2 & 30.7 & 34.9 & 67.2 & 70.0 \\
& Real-IAD Variety S3 & 100 & 76.5 & 94.9 & 93.7 & 89.2 & 27.0 & 32.2 & 66.5 & 68.6 \\
& Real-IAD Variety & 160 & 73.3 & 94.2 & 93.5 & 88.3 & 24.0 & 29.4 & 64.5 & 66.7 \\
\rowcolor[gray]{0.95}
\multicolumn{3}{c}{$\Delta \downarrow$} & 9.5 & 2.5 & 0.9 & 1.8 & 12.7 & 9.7 & 4.5 & 5.9 \\

\midrule

\multirow{5}{*}{MambaAD~\cite{he2024mambaad}}
& Real-IAD Variety S1 & 30 & 87.4 & 97.5 & 95.0 & 91.5 & 40.3 & 43.2 & 75.0 & 75.7 \\
& Real-IAD Variety S2 & 60 & 76.2 & 94.9 & 93.8 & 89.0 & 25.9 & 31.5 & 68.1 & 68.5 \\
& Real-IAD Variety S3 & 100 & 72.5 & 93.7 & 93.6 & 88.5 & 21.5 & 27.5 & 66.5 & 66.3 \\
& Real-IAD Variety & 160 & 81.7 & 96.2 & 93.9 & 91.2 & 33.1 & 37.7 & 73.5 & 72.5 \\
\rowcolor[gray]{0.95}
\multicolumn{3}{c}{$\Delta \downarrow$} & 14.9 & 3.8 & 1.4 & 3.0 & 18.8 & 15.7 & 8.5 & 9.4 \\

\midrule

\multirow{5}{*}{LGC~\cite{fan2025salvaging}}
& Real-IAD Variety S1 & 30 & 84.1 & 96.7 & 94.8 & 91.0 & 36.1 & 39.9 & 73.7 & 73.8 \\
& Real-IAD Variety S2 & 60 & 82.7 & 96.4 & 94.1 & 91.9 & 34.5 & 39.9 & 76.6 & 73.7 \\
& Real-IAD Variety S3 & 100 & 82.4 & 96.2 & 94.3 & 91.8 & 32.8 & 37.9 & 75.0 & 72.9 \\
& Real-IAD Variety & 160 & 81.0 & 96.0 & 94.0 & 91.7 & 31.9 & 37.0 & 74.6 & 72.3 \\
\rowcolor[gray]{0.95}
\multicolumn{3}{c}{$\Delta \downarrow$} & 3.1 & 0.7 & 0.8 & 0.9 & 4.2 & 2.9 & 2.9 & 1.4 \\

\midrule

\multirow{5}{*}{Dinomaly~\cite{guo2025dinomaly}}
& Real-IAD Variety S1 & 30 & 91.4 & 98.4 & 95.8 & 92.7 & 56.9 & 56.1 & 79.8 & 81.6 \\
& Real-IAD Variety S2 & 60 & 89.7 & 98.1 & 95.2 & 92.6 & 53.1 & 53.7 & 79.7 & 80.3 \\
& Real-IAD Variety S3 & 100 & 87.9 & 97.7 & 94.9 & 92.3 & 48.3 & 50.1 & 78.2 & 78.5 \\
& Real-IAD Variety & 160 & 85.4 & 97.2 & 94.5 & 91.5 & 42.8 & 45.8 & 75.6 & 76.1 \\
\rowcolor[gray]{0.95}
\multicolumn{3}{c}{$\Delta \downarrow$} & 6.0 & 1.2 & 1.3 & 1.2 & 14.1 & 10.3 & 4.2 & 5.5 \\

\midrule

\multirow{5}{*}{Dinomaly+~\cite{guo2025dinomaly}}
& Real-IAD Variety S1 & 30 & 92.4 & 98.6 & 96.0 & 92.7 & 59.2 & 56.1 & 79.3 & 82.0 \\
& Real-IAD Variety S2 & 60 & 90.8 & 98.3 & 95.3 & 92.9 & 56.3 & 54.4 & 79.6 & 81.1 \\
& Real-IAD Variety S3 & 100 & 89.4 & 97.9 & 94.7 & 92.4 & 53.6 & 52.5 & 78.4 & 79.8 \\
& Real-IAD Variety & 160 & 87.1 & 97.3 & 94.0 & 91.9 & 49.7 & 49.2 & 76.4 & 77.9 \\
\rowcolor[gray]{0.95}
\multicolumn{3}{c}{$\Delta \downarrow$} & 5.3 & 1.3 & 2.0 & 1.0 & 9.5 & 6.9 & 3.2 & 4.1 \\

\midrule

\multirow{5}{*}{\textbf{DPDiff-AD}}
& Real-IAD Variety S1 & 30 & 93.5 & 98.7 & 96.5 & 93.9 & 49.6 & 50.5 & 77.7 & 80.1 \\
& Real-IAD Variety S2 & 60 & 93.2 & 98.7 & 96.2 & 94.4 & 50.1 & 51.3 & 79.0 & 80.4 \\
& Real-IAD Variety S3 & 100 & 92.4 & 98.4 & 95.9 & 94.6 & 50.7 & 51.3 & 79.8 & 80.4 \\
& Real-IAD Variety & 160 & 92.4 & 98.5 & 95.8 & 94.8 & 50.1 & 50.7 & 80.1 & 80.4 \\
\rowcolor[gray]{0.95}
\multicolumn{3}{c}{$\Delta \downarrow$} & \textbf{1.1} &\textbf{ 0.3} & \textbf{0.7} & \textbf{0.9} & \textbf{1.1} & \textbf{0.8}& \textbf{2.4} & \textbf{0.3} \\

\bottomrule
\end{tabular}
}

\end{table*}

\subsubsection{Scalability Analysis under Increasing Category Scales}

To further examine the robustness of MUAD methods under category expansion, we conduct scalability evaluation using the official S1/S2/S3/Full protocol of Real-IAD Variety. This protocol progressively increases the category scale from 30 to 160 classes, providing a challenging benchmark for assessing whether a unified anomaly detection model can maintain stable performance under increasingly diverse normal distributions.

As shown in Table~\ref{tab:Scalability evaluation}, most existing MUAD methods exhibit clear performance fluctuations or degradation as the category space grows, indicating limited robustness to increasingly heterogeneous normal distributions. To quantify the sensitivity of each method to category expansion, we report $\Delta$, defined as the absolute difference between the maximum and minimum values across the four category-scale settings. For example, UniAD shows an 11.0-point variation in image-level AUROC, while recent competitive methods such as MambaAD and Dinomaly+ exhibit variations of 14.9 and 5.3 points, respectively. Similar instability is also observed across several image-level and pixel-level metrics, suggesting that category expansion remains challenging even for strong MUAD models.

In contrast, DPDiff-AD maintains stable performance throughout the scaling process. Across the four category-scale settings, the image-level AUROC, AP, and F$_1$-max show only marginal variations of 1.1, 0.3, and 0.7 points, respectively. For pixel-level localization, DPDiff-AD also remains stable, with variations of only 0.9, 1.1, 0.8, and 2.4 points in AUROC, AP, F$_1$-max, and AUPRO, respectively. Notably, the overall mAD variation is only 0.3, which is substantially smaller than that of the previous state-of-the-art method Dinomaly+ (4.1). These results indicate that DPDiff-AD is less sensitive to category expansion and preserves stable anomaly detection and localization performance even under the large-scale 160-category setting.

We attribute this robustness to the proposed dual-scale prototype-conditioned diffusion framework. Specifically, local prototypes stabilize patch-level semantic structures, while global prototypes regularize the overall feature distribution through optimal transport, jointly reducing representation fragmentation and inter-category interference under category expansion. Furthermore, the prototype-conditioned diffusion reconstruction module incorporates these dual-scale prototypes into the denoising process through prototype-aware attention, enabling the diffusion model to selectively attend to representative normal patterns during feature reconstruction. In this way, the model can exploit structured prototype priors to guide the reconstruction of complex multi-modal normal distributions along a more coherent normal manifold. As a result, DPDiff-AD maintains stable normality modeling and anomaly discrimination even in the large-scale 160-category setting.

\begin{table*}[t]
\centering
\caption{Quantitative MUAD performance on four widely used anomaly detection benchmarks.}
\label{tab:MVAD_four_data}
\resizebox{0.95\textwidth}{!}{
\begin{tabular}{l|ccccccc c|ccccccc c}
\toprule
\multirow{3}{*}{Method} 
& \multicolumn{8}{c|}{\textbf{MVTec-AD} (15 classes)} 
& \multicolumn{8}{c}{\textbf{VisA} (12 classes)} \\

& \multicolumn{3}{c}{Image-level} 
& \multicolumn{4}{c}{Pixel-level}
& \multirow{2}{*}{mAD}
& \multicolumn{3}{c}{Image-level} 
& \multicolumn{4}{c}{Pixel-level}
& \multirow{2}{*}{mAD} \\

\cmidrule(r){2-4} \cmidrule(lr){5-8}
\cmidrule(r){10-12} \cmidrule(l){13-16}

& AUROC & AP & F$_1$-max 
& AUROC & AP & F$_1$-max & AUPRO
&
& AUROC & AP & F$_1$-max
& AUROC & AP & F$_1$-max & AUPRO
& \\

\midrule

RD4AD~\cite{deng2022anomaly} & 94.6 & 96.5 & 95.2 & 96.1 & 48.6 & 53.8 & 91.1 & 82.3
& 92.4 & 92.4 & 89.6 & 98.1 & 38.0 & 42.6 & 91.8 & 77.8 \\

SimpleNet~\cite{liu2023simplenet} & 95.3 & 98.4 & 95.8 & 96.9 & 45.9 & 49.7 & 86.5 & 81.2
& 87.2 & 87.0 & 81.8 & 96.8 & 34.7 & 37.8 & 81.4 & 72.4 \\

DeSTSeg~\cite{zhang2023destseg} & 89.2 & 95.5 & 91.6 & 93.1 & 54.3 & 50.9 & 64.8 & 77.1
& 88.9 & 89.0 & 85.2 & 96.1 & 39.6 & 43.4 & 67.4 & 72.8 \\

UniAD~\cite{you2022unified} & 96.5 & 98.8 & 96.2 & 96.8 & 43.4 & 49.5 & 90.7 & 81.7
& 91.4 & 93.3 & 87.5 & 98.5 & 35.3 & 40.2 & 89.0 & 76.5 \\

ReContrast~\cite{guo2023recontrast} & 98.3 & 99.4 & 97.6 & 97.1 & 60.2 & 61.5 & 93.2 & 86.8
& 95.5 & 96.4 & 92.0 & 98.5 & 47.9 & 50.6 & 91.9 & 81.8 \\

DiAD~\cite{he2024diffusion} & 97.2 & 99.0 & 96.5 & 96.8 & 52.6 & 55.5 & 90.7 & 84.0
& 86.8 & 88.3 & 85.1 & 96.0 & 26.1 & 33.0 & 75.2 & 70.1 \\

HVQ-Trans~\cite{lu2023hierarchical} & 98.0 & 99.5 & 97.5 & 97.3 & 48.2 & 53.3 & 91.4 & 83.6
& 93.2 & 92.8 & 87.6 & 98.7 & 35.0 & 39.6 & 86.3 & 76.2 \\

ViTAD~\cite{zhang2023exploring} & 98.3 & 99.4 & 97.3 & 97.7 & 55.3 & 58.7 & 91.4 & 85.4
& 90.5 & 91.7 & 86.3 & 98.2 & 36.6 & 41.1 & 85.1 & 75.6 \\

MambaAD~\cite{he2024mambaad} & 98.6 & 99.6 & 97.8 & 97.7 & 56.3 & 59.2 & 93.1 & 86.0
& 94.3 & 94.5 & 89.4 & 98.5 & 39.4 & 44.0 & 91.0 & 78.7 \\

\rowcolor[gray]{0.95}
\textbf{DPDiff-AD} 
& \textbf{99.5} & \textbf{99.9} & \textbf{99.1}
& \textbf{98.7} & \textbf{65.6} & \textbf{64.8} & \textbf{95.9} & \textbf{89.1}
& \textbf{97.7} & \textbf{98.5} & \textbf{95.5}
& \textbf{99.1} & \textbf{49.2} & \textbf{51.5} & \textbf{94.5} & \textbf{83.7} \\

\hline\hline

\multirow{3}{*}{Method} 
& \multicolumn{8}{c|}{\textbf{MPDD} (6 classes)} 
& \multicolumn{8}{c}{\textbf{Real-IAD} (30 classes)} \\

& \multicolumn{3}{c}{Image-level} 
& \multicolumn{4}{c}{Pixel-level}
& \multirow{2}{*}{mAD}
& \multicolumn{3}{c}{Image-level} 
& \multicolumn{4}{c}{Pixel-level}
& \multirow{2}{*}{mAD} \\

\cmidrule(r){2-4} \cmidrule(lr){5-8}
\cmidrule(r){10-12} \cmidrule(l){13-16}

& AUROC & AP & F$_1$-max 
& AUROC & AP & F$_1$-max & AUPRO
&
& AUROC & AP & F$_1$-max 
& AUROC & AP & F$_1$-max & AUPRO
& \\

\midrule

RD4AD~\cite{deng2022anomaly} & 90.3 & 92.8 & 90.5 & 98.3 & 39.6 & 40.6 & 95.2 & 78.2
& 82.4 & 79.0 & 73.9 & 97.3 & 25.0 & 32.7 & 89.6 & 68.6 \\

SimpleNet~\cite{liu2023simplenet} & 90.6 & 94.1 & 89.7 & 97.1 & 33.6 & 35.7 & 90.0 & 75.8
& 57.2 & 53.4 & 61.5 & 75.7 & 2.8 & 6.5 & 39.0 & 42.3 \\

DeSTSeg~\cite{zhang2023destseg} & 92.6 & 91.8 & 92.8 & 90.8 & 30.6 & 32.9 & 78.3 & 72.8
& 82.3 & 79.2 & 73.2 & 94.6 & \textbf{37.9} & \textbf{41.7} & 40.6 & 64.2 \\

UniAD~\cite{you2022unified} & 80.1 & 83.2 & 85.1 & 95.4 & 19.0 & 25.6 & 83.8 & 67.5
& 83.0 & 80.9 & 74.3 & 97.3 & 21.1 & 29.2 & 86.7 & 67.5 \\

ReContrast~\cite{guo2023recontrast} & 90.9 & 94.5 & 91.0 & \textbf{98.8} & 46.8 & \textbf{48.9} & \textbf{95.6} & 80.9
& 86.4 & 84.2 & 77.4 & 97.8 & 31.6 & 38.2 & 91.8 & 72.5 \\

DiAD~\cite{he2024diffusion} & 85.8 & 89.2 & 86.5 & 91.4 & 15.3 & 19.2 & 66.1 & 64.8
& 75.6 & 66.4 & 69.9 & 88.0 & 2.9 & 7.1 & 58.1 & 52.6 \\

HVQ-Trans~\cite{lu2023hierarchical} & 86.5 & 88.1 & 85.8 & 96.7 & 27.6 & 31.4 & 86.9 & 71.9
& 86.6 & 84.9 & 79.4 & 98.0 & 27.6 & 34.4 & 88.7 & 71.4 \\

ViTAD~\cite{zhang2023exploring} & 87.4 & 90.8 & 87.0 & 97.8 & 44.1 & 46.4 & 95.3 & 78.4
& 82.7 & 80.2 & 73.7 & 97.2 & 24.3 & 32.3 & 84.8 & 67.9 \\

MambaAD~\cite{he2024mambaad} & 89.2 & 93.1 & 90.3 & 97.7 & 33.5 & 38.6 & 92.8 & 76.5
& 86.3 & 84.6 & 77.0 & 98.5 & 33.0 & 38.7 & 90.5 & 72.7 \\

\rowcolor[gray]{0.95}
\textbf{DPDiff-AD} 
& \textbf{97.0} & \textbf{98.5} & \textbf{96.5}
& 98.5 & \textbf{46.9} & 48.1 & 95.4 & \textbf{83.0}
& \textbf{88.9} & \textbf{85.4} & \textbf{80.0}
& \textbf{99.1} & 31.6 & 37.0 & \textbf{93.1} & \textbf{73.6} \\

\bottomrule
\end{tabular}
}
\end{table*}

\subsection{Evaluations on Widely Used Anomaly Detection Benchmarks}
Beyond the large-scale Real-IAD Variety benchmark, we further evaluate DPDiff-AD on four widely used industrial anomaly detection benchmarks, namely MVTec-AD, VisA, MPDD, and Real-IAD, to assess its generalization capability across diverse industrial scenarios. The quantitative results are reported in Table~\ref{tab:MVAD_four_data}.

On MVTec-AD, DPDiff-AD achieves highly competitive performance across both image-level anomaly detection and pixel-level anomaly localization metrics. Compared with the strongest competing method, ReContrast, DPDiff-AD improves the overall mAD by 2.3 points and achieves higher scores on all reported metrics. Specifically, it surpasses ReContrast by 1.2 points in image-level AUROC (99.5\% vs. 98.3\%) and 1.6 points in pixel-level AUROC (98.7\% vs. 97.1\%), while also achieving 65.6\% pixel-level AP and 95.9\% AUPRO. On VisA, DPDiff-AD also achieves the best overall performance, exceeding the previous best method, ReContrast, by 2.2 points in image-level AUROC, 0.6 points in pixel-level AUROC, and 2.6 points in AUPRO. These results demonstrate the effectiveness of DPDiff-AD on widely used industrial anomaly detection benchmarks.

DPDiff-AD also achieves strong results on MPDD and Real-IAD. On MPDD, DPDiff-AD obtains the best overall mAD of 83.0\%, outperforming the strongest competing method, ReContrast, by 2.1 points. It also achieves the best image-level AUROC, AP, and F$_1$-max of 97.0\%, 98.5\%, and 96.5\%, respectively, as well as the best pixel-level AP of 46.9\%. Although ReContrast obtains slightly higher pixel-level AUROC, F$_1$-max, and AUPRO, DPDiff-AD shows better overall performance. On Real-IAD, DPDiff-AD obtains the best image-level AUROC, AP, and F$_1$-max of 88.9\%, 85.4\%, and 80.0\%, respectively. It also achieves the best pixel-level AUROC of 99.1\% and the highest AUPRO of 93.1\%, leading to the best overall mAD of 73.6\%.

Overall, these results demonstrate that DPDiff-AD not only scales effectively to large and diverse category spaces, but also generalizes well across widely used industrial anomaly detection benchmarks, delivering robust and balanced performance on datasets with different scales and characteristics.

\begin{figure*}[t!]\vspace{0mm}
		\centering
		\includegraphics[width=125mm]{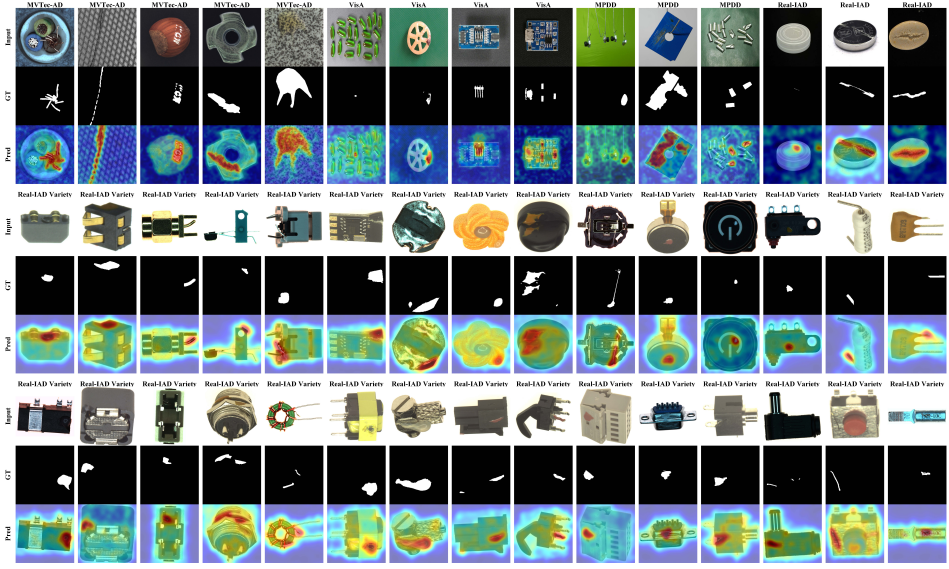} \vspace{0mm}
\caption{Qualitative anomaly localization results of  DPDiff-AD on five datasets. For each example, we present the input image, the ground-truth mask (GT), and the predicted anomaly map (Pred).}
	\label{fig.Qualitative results}
	\end{figure*}

\subsection{Qualitative Analysis of Anomaly Localization}

To provide intuitive evidence of the effectiveness of  DPDiff-AD, we visualize the anomaly localization results in Fig.~\ref{fig.Qualitative results}.  DPDiff-AD consistently produces anomaly maps that closely correspond to the ground-truth regions across diverse defect patterns. On commonly used benchmarks, including MVTec-AD, VisA, MPDD, and Real-IAD, the proposed method accurately highlights anomalous regions while effectively suppressing spurious responses in normal areas. More importantly,  DPDiff-AD remains highly effective on the large-scale Real-IAD Variety dataset, where substantially increased category diversity and visual complexity make anomaly localization considerably more challenging. Even under such highly heterogeneous settings,  DPDiff-AD still yields concentrated responses on defective regions with limited background interference, demonstrating its robustness in large-scale industrial scenarios. Additional qualitative visualizations of DPDiff-AD across different datasets and categories are provided in Appendix~\ref{append_vision}.

\begin{figure*}[t!]\vspace{0mm}
		\centering
		\includegraphics[width=125mm]{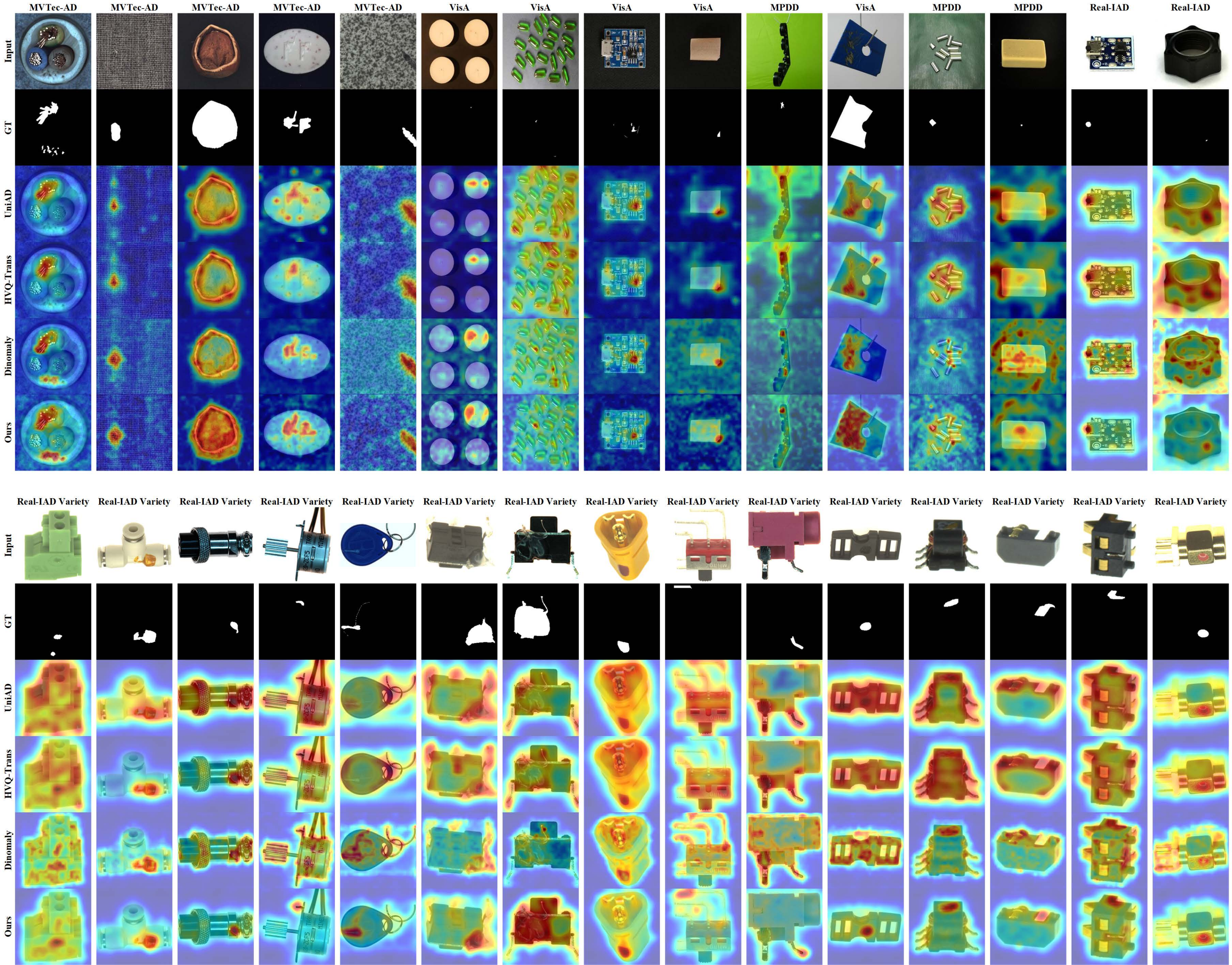} \vspace{0mm}
\caption{Qualitative anomaly localization comparisons on commonly used benchmarks and Real-IAD Variety. We compare DPDiff-AD with UniAD, HVQ-Transformer, and Dinomaly using input images, ground-truth masks (GT), and predicted anomaly maps.}
        \label{fig.Qualitative contras}
	\end{figure*}

Furthermore, as shown in Fig.~\ref{fig.Qualitative contras}, we compare DPDiff-AD with representative methods on both benchmarks and the large-scale Real-IAD Variety dataset. For a fair comparison, all anomaly maps are visualized using the same heatmap post-processing protocol as UniAD~\cite{you2022unified}, since different normalization or smoothing strategies may lead to visually different localization effects. Under this unified visualization protocol, DPDiff-AD produces anomaly maps that are well aligned with the ground-truth regions, exhibiting clearer boundaries and reduced background noise on commonly used anomaly detection benchmarks. Compared with Dinomaly, it shows a slight advantage, while significantly outperforming UniAD and HVQ-Transformer. More importantly, on the large-scale Real-IAD Variety dataset, where category cardinality increases and the normal distribution becomes highly heterogeneous, the limitations of existing unified models become increasingly evident. UniAD and HVQ-Transformer exhibit severe localization failures in such scenarios, while Dinomaly also suffers from noticeable performance degradation, often producing incomplete or overly diffuse anomaly responses. In contrast, DPDiff-AD remains robust and consistently delivers accurate anomaly localization across diverse categories and complex defect patterns. These qualitative results further validate the effectiveness of the proposed prototype-conditioned diffusion framework, showing that dual-scale prototype priors guide the diffusion reconstruction process toward more coherent normal reconstructions and reliable anomaly responses.

\begin{table*}[t]
\centering
\footnotesize
\setlength{\tabcolsep}{4pt}
\renewcommand{\arraystretch}{1.1}

\caption{Inference speed (FPS) of different models on various datasets with batch size 64.}
\label{table:fps}
\resizebox{0.95\textwidth}{!}{
\begin{tabular}{c|ccc|ccc|c}
\toprule
\multirow{2}{*}{\textbf{Dataset}} 
& \multicolumn{3}{c|}{\textbf{Transformer-based}} 
& \multicolumn{3}{c|}{\textbf{Diffusion-based}} 
& \multirow{2}{*}{\textbf{DPDiff-AD}} \\
\cmidrule(lr){2-4}
\cmidrule(lr){5-7}
& HVQ-Trans~\cite{lu2023hierarchical} & UniAD~\cite{you2022unified} & Dinomaly~\cite{guo2025dinomaly}
& DDAD~\cite{mousakhan2024anomaly} & DiAD~\cite{he2024diffusion} & TransFusion~\cite{fuvcka2024transfusion}
&  \\
\midrule
MVTec-AD            & 356.546 & 338.489 & 128.738 & 10.653 & 4.086 & 4.024 & 101.426 \\
VisA                & 317.148 & 291.174 & 126.808  & 10.622 & 3.892 & 4.006 & 101.298 \\
MPDD                & 395.062 & 371.451 & 130.799  & 10.804 & 3.919 & 3.999 & 101.170 \\
Real-IAD            & 397.762 & 369.516 & 127.095 & 10.739 & 4.112 & 3.987 & 102.253 \\
Real-IAD Variety    & 345.759 & 337.731 & 120.937 & 10.634 &3.898 & 4.018 & 100.016\\
Average FPS  & 362.455 & 341.672& 126.875& 10.690&3.981& 4.007& 101.233\\
\bottomrule
\end{tabular}
}
\end{table*}
\subsection{Inference Speed Comparison}
As shown in Table~\ref{table:fps},  DPDiff-AD achieves a favorable balance between inference efficiency and detection performance. Across five datasets, it runs at approximately 101 FPS on average, significantly outperforming existing diffusion-based methods in terms of inference speed, with 9.5$\times$, 25.4$\times$, and 25.3$\times$ speedups over DDAD, DiAD, and TransFusion, respectively. Compared with transformer-based methods,  DPDiff-AD remains competitive in efficiency, achieving an inference speed close to that of Dinomaly. Although HVQ-Trans and UniAD exhibit higher FPS, their detection performance is notably inferior, as evidenced by the lower image-level and pixel-level AUROC reported in Table~\ref{Real-IAD-Variety} and Table~\ref{tab:MVAD_four_data}. These results demonstrate that DPDiff-AD effectively alleviates the efficiency bottleneck of diffusion-based anomaly detection while preserving strong detection capability, thereby offering a more practical accuracy--efficiency trade-off for real-world industrial applications. This efficiency is further analyzed in Sec.~\ref{ablation_infer}, where we show that DPDiff-AD can maintain stable performance with only a few DDIM~\cite{song2020denoising} sampling steps.

\subsection{Ablation Studies and Analysis}

Due to the substantially larger scale of Real-IAD Variety and the high computational cost of repeatedly training multiple ablation variants on such a large-scale benchmark, we conduct the ablation studies on the widely used MVTec-AD dataset. This choice follows the standard evaluation practice in industrial anomaly detection, where MVTec-AD is commonly used for controlled ablation analysis because of its representative defect categories, high-quality pixel-level annotations, and moderate experimental scale. Specifically, we evaluate the contributions of the local prototype (L-Proto), global prototype (G-Proto), and Prototype-Conditioned Diffusion Reconstruction (Diffusion). As shown in Table~\ref{tab:ablation}, directly using the local or global prototypes for similarity-based anomaly detection leads to limited performance. This indicates that the prototypes are not intended to serve as standalone feature templates for direct matching. Instead, their effectiveness is fully revealed when they are integrated into the diffusion reconstruction process. Compared with the prototype-only variants, the diffusion-only baseline achieves clearly better results and provides a strong reconstruction foundation. Further incorporating either local or global prototypes into the diffusion module consistently improves the performance, suggesting that the prototype-aware attention mechanism can effectively exploit prototype information during denoising-based feature reconstruction. Combining both prototype modules yields the best overall results, further demonstrating the complementarity between local structural priors and global distributional priors.
\begin{table*}[t]
\centering
\caption{Ablation study of the proposed components on MVTec-AD. L-Proto, G-Proto, and Diffusion denote the local prototype, global prototype, and Prototype-Conditioned Diffusion Reconstruction modules, respectively.}
\label{tab:ablation}
\setlength{\tabcolsep}{5pt}
\renewcommand{\arraystretch}{1.05}
\footnotesize
\resizebox{0.95\textwidth}{!}{
\begin{tabular}{c c c ccc cccc c}
\toprule
\multicolumn{3}{c}{Module} & \multicolumn{3}{c}{Image-level} & \multicolumn{4}{c}{Pixel-level} & \multirow{2}{*}{mAD} \\
\cmidrule(lr){1-3} \cmidrule(lr){4-6} \cmidrule(lr){7-10}
L-Proto & G-Proto & Diffusion & AUROC & AP & F$_1$-max & AUROC & AP & F$_1$-max & AUPRO &  \\
\midrule
$\checkmark$  & $\times$      & $\times$ & 72.7 & 87.7 & 87.3 & 83.7 & 27.1 & 31.9 & 77.6 &66.9 \\
$\times$      & $\checkmark$  & $\times$ & 71.7 & 84.7 & 84.9 & 79.2 & 18.8 & 24.2 & 61.3 & 60.7 \\
$\times$      & $\times$      & $\checkmark$ & 98.9 & 99.6 & 98.3 & 98.0 & 62.6 & 63.6 & 94.8 & 88.0 \\
$\checkmark$  & $\times$      & $\checkmark$ & 99.2 & 99.8 & 98.8 & 98.5 & 64.6 & 64.7 & 95.4 & 88.7 \\
$\times$      & $\checkmark$  & $\checkmark$ & 99.1 & 99.7 & 98.6 & 98.4 & 64.5 & \textbf{64.8} & 95.4 & 88.6 \\
\rowcolor[gray]{0.95}
$\checkmark$  & $\checkmark$  & $\checkmark$ & \textbf{99.5} & \textbf{99.9} & \textbf{99.1} &\textbf{ 98.7} & \textbf{65.6} & \textbf{64.8} & \textbf{95.9} & \textbf{89.1} \\
\bottomrule
\end{tabular}
}
\end{table*}

\begin{figure*}[t!]\vspace{0mm}
		\centering
		\includegraphics[width=125mm]{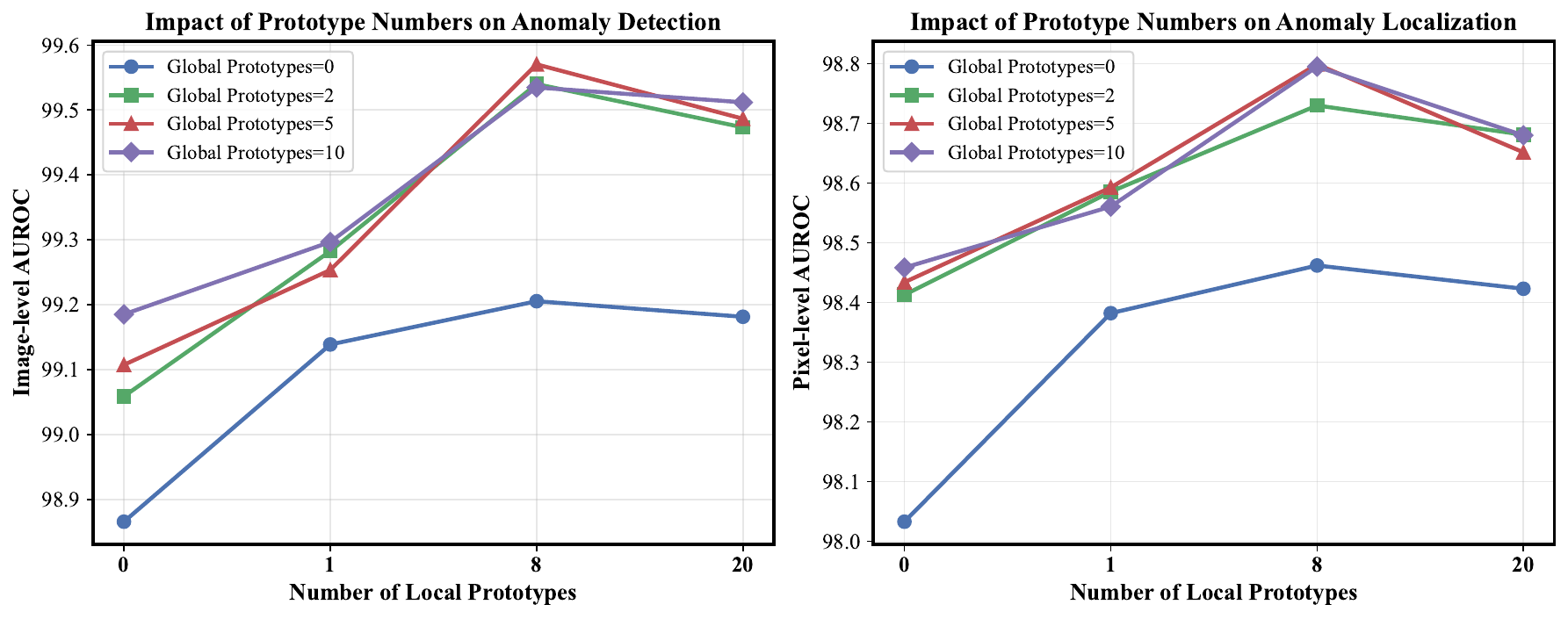} \vspace{0mm}
\caption{Impact of local and global prototypes on anomaly detection and localization performance.}
        \label{fig.ablation_global_local}
	\end{figure*}

\subsubsection{Impact of Prototype Numbers}

As shown in Fig.~\ref{fig.ablation_global_local}, both local and global prototypes play important roles in anomaly detection and localization. Compared with the setting without global prototypes, introducing global prototypes clearly improves the overall performance, indicating their effectiveness in capturing the global distribution of normal data. In addition, increasing the number of local prototypes generally improves performance up to a moderate scale, as it enables the model to capture more fine-grained structural patterns. However, further increasing the number of local prototypes to 20 leads to a slight performance decrease compared with the setting using 8 local prototypes. This suggests that excessive prototypes may introduce redundant normal patterns and unnecessary complexity, thereby weakening generalization. Overall, these results highlight the complementary roles of local and global prototypes and the importance of using a moderate prototype scale.

\subsubsection{Ablation on Diffusion Inference and Sampling Steps} \label{ablation_infer}

\begin{figure*}[t!]\vspace{0mm}
		\centering
		\includegraphics[width=125mm]{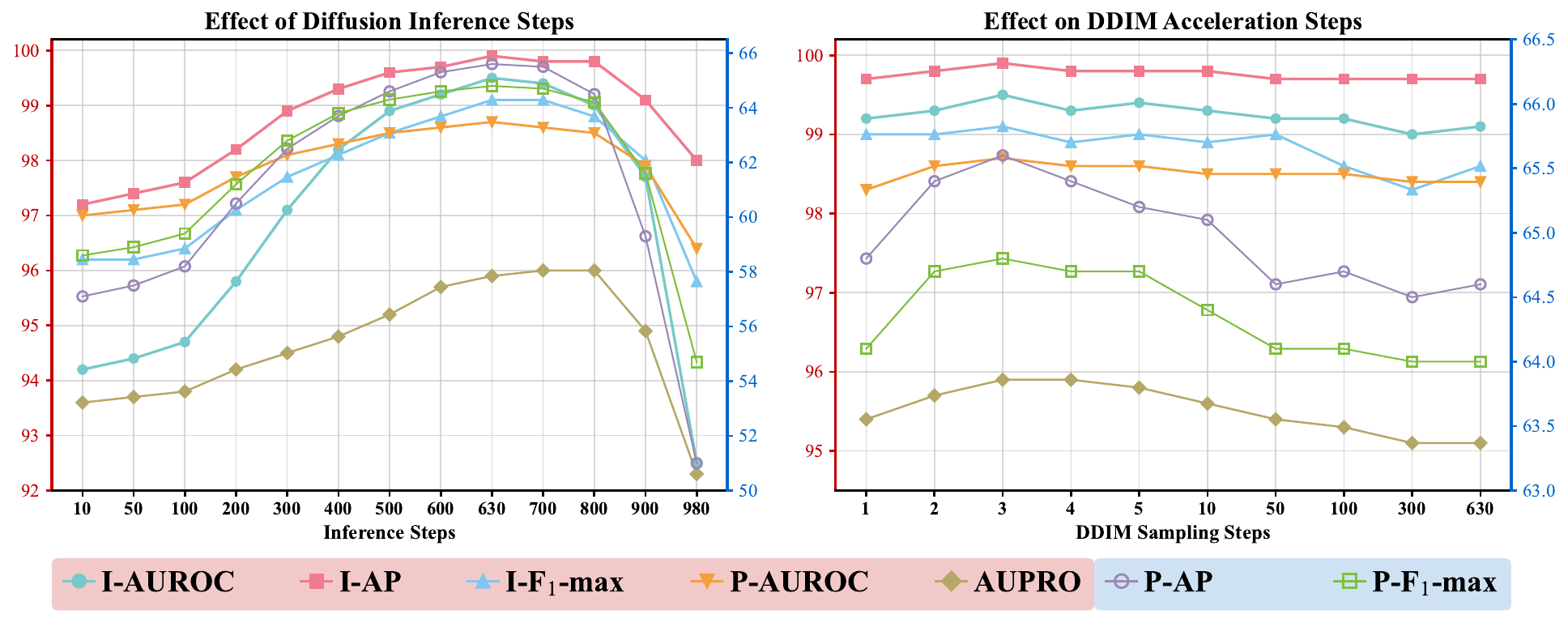} \vspace{0mm}
\caption{Effect of diffusion inference steps and DDIM sampling steps on anomaly detection and localization performance. Pixel-level AP and F$_1$-max are plotted on the right blue axis, while all other metrics share the left red axis.}
        \label{fig.ablation_inference_ddim}
	\end{figure*}

As shown in Fig.~\ref{fig.ablation_inference_ddim}, we investigate the effects of diffusion inference steps and DDIM sampling steps on anomaly detection and localization performance. For diffusion inference steps, the performance improves as the number of steps increases from a small value and reaches its optimum within a moderate range. However, when the number of steps becomes excessively large, the performance gradually decreases. This degradation is mainly because excessive noise levels may obscure the original semantic information, making effective reconstruction more difficult.

For DDIM sampling steps, near-optimal performance can be achieved with only a few steps, and the best overall performance is obtained with 3 steps. This efficiency can be attributed to the hierarchical feature aggregation mechanism and the explicit guidance of dual prototypes during the denoising process. Specifically, prototype-aware attention enables the diffusion model to effectively exploit these prototype priors, allowing it to recover normal structures and maintain strong anomaly discrimination even with very few sampling steps. Such a small number of sampling steps substantially reduces the diffusion inference cost and detection time, making the reconstruction process both efficient and stable. As the number of DDIM sampling steps further increases, the performance remains nearly unchanged, indicating that the introduced structural priors already provide sufficient guidance with a small number of sampling steps.

\subsubsection{t-SNE Visualization of Feature Distributions}

\begin{figure*}[t!]\vspace{0mm}
		\centering
		\includegraphics[width=125mm]{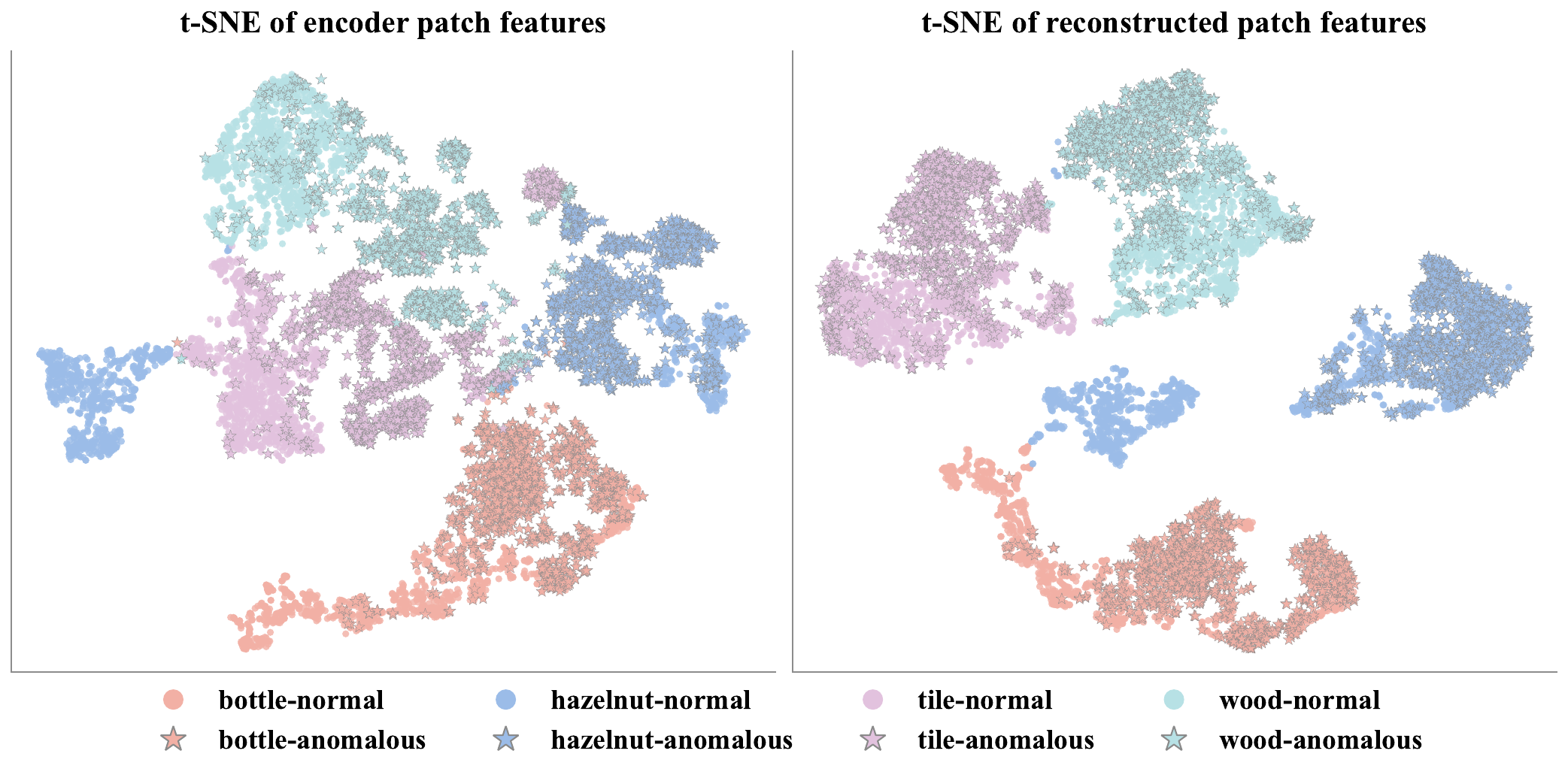} \vspace{0mm}
\caption{t-SNE visualization of patch features before and after reconstruction. The left panel shows the encoder features, while the right panel shows the reconstructed features. $\CIRCLE$ and $\bigstar$ denote normal and anomalous samples, respectively.}
        \label{fig.ablation_tsne_two_panel}
	\end{figure*}

As shown in Fig.~\ref{fig.ablation_tsne_two_panel}, we visualize the feature distributions before and after reconstruction using t-SNE. In the encoder feature space, normal and anomalous patches exhibit separable distributions. After prototype-guided diffusion reconstruction, anomalous patches are pulled toward the corresponding normal clusters. This observation indicates that, under the guidance of learned prototype priors, the reconstruction process tends to recover anomalous regions as normal-like patterns. Consequently, compared with normal regions, anomalous regions produce larger discrepancies between the encoder features and the reconstructed features, enabling effective anomaly discrimination based on reconstruction loss.

\subsubsection{Analysis of Prototype Similarity}
\begin{figure*}[t!]\vspace{0mm}
		\centering
		\includegraphics[width=90mm]{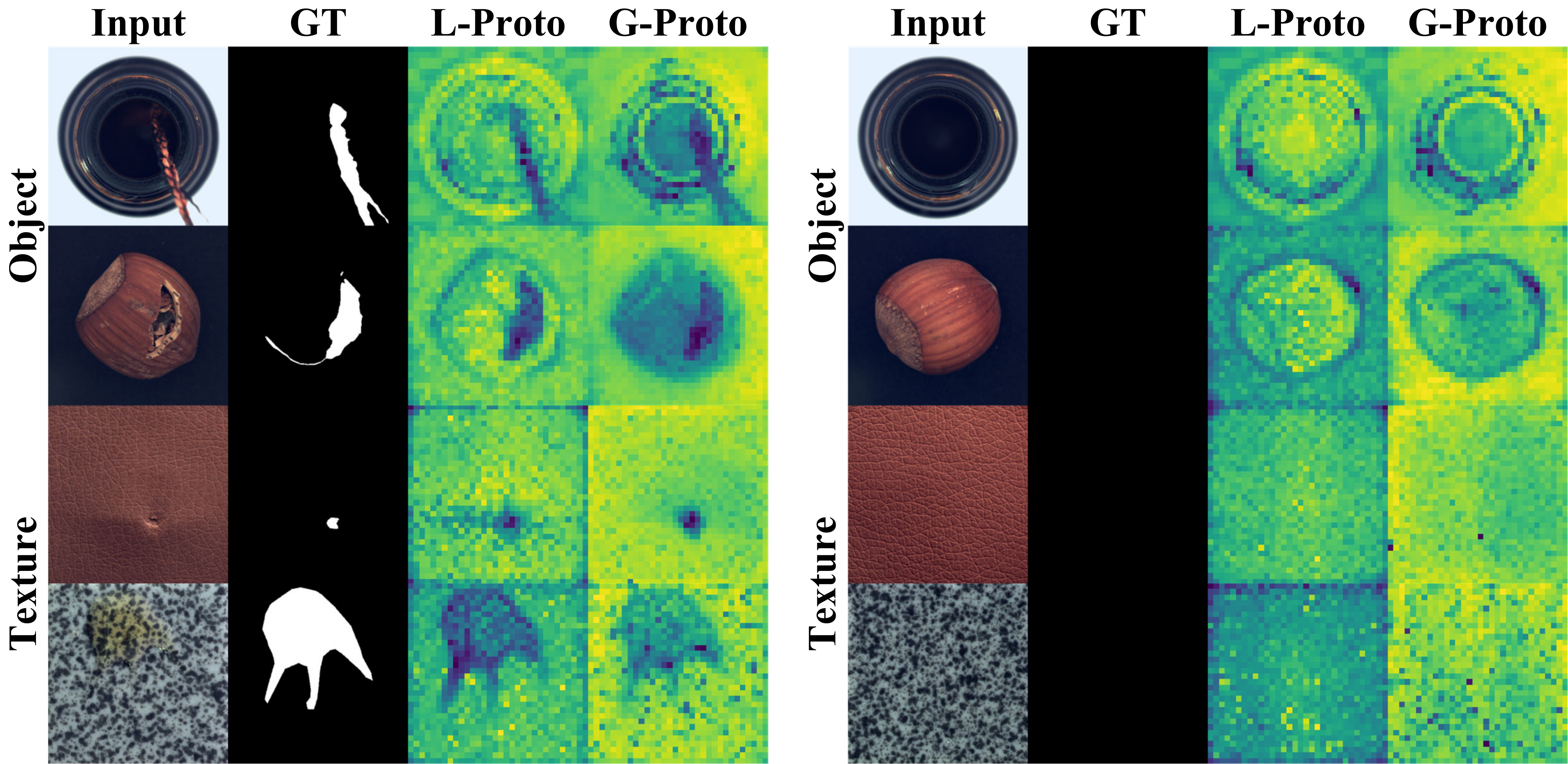} \vspace{0mm}
\caption{Visualization of similarity maps for local prototypes (L-Proto) and global prototypes (G-Proto). The left and right panels show anomalous and normal samples, respectively. Both prototypes mainly respond to normal regions, while L-Proto and G-Proto capture local details and global structures, respectively.}
        \label{fig.ablation_see_prorotype}
	\end{figure*}

To further understand what the prototypes learn, we visualize the similarity maps of local prototypes (L-proto) and global prototypes (G-proto) on both anomalous and normal samples, as shown in Fig.~\ref{fig.ablation_see_prorotype}. For anomalous samples (left), both types of prototypes exhibit high similarity to normal regions but low similarity to anomalous regions, indicating that they primarily capture normal patterns rather than anomalous structures. For normal samples (right), both prototypes respond consistently to semantically meaningful normal regions, further confirming that the learned prototypes serve as reliable normality priors. Moreover, local prototypes tend to focus more on fine-grained local structures, such as boundaries, textures, and defect-adjacent details, while global prototypes place more emphasis on holistic object shapes, semantic layouts, and global normal patterns. This difference suggests that local and global prototypes provide complementary cues for modeling normal data, jointly guiding the reconstruction process toward the normal manifold.

\subsubsection{Ablation on Input Resolution}
\begin{table*}[t]
\centering
\caption{Effect of input resolution on anomaly detection and localization performance on MVTec-AD.}
\label{tab:input_size}
\setlength{\tabcolsep}{3pt}
\renewcommand{\arraystretch}{0.95}
\footnotesize
\resizebox{0.95\columnwidth}{!}{
\begin{tabular}{c ccc cccc c}
\toprule
\multirow{2}{*}{Input Size} &
\multicolumn{3}{c}{Image-level} &
\multicolumn{4}{c}{Pixel-level} &
\multirow{2}{*}{mAD} \\
\cmidrule(lr){2-4}
\cmidrule(lr){5-8}
& AUROC & AP & F$_1$-max &
AUROC & AP & F$_1$-max & AUPRO &
\\
\midrule
$224^2$ & 90.0 & 99.7 & 98.5 & 98.1 & 55.1 & 57.4 & 92.6 & 84.5 \\
$336^2$ & 99.4 & 99.8 & 99.0 & 98.5 & 61.6 & 62.3 & 95.1 & 88.0 \\
$448^2$ & \textbf{99.5} & \textbf{99.9} & \textbf{99.1} & 98.7 & 65.6 & 64.8 & 95.9 & 89.1 \\
$560^2$ &\textbf{99.5} & \textbf{99.9} & \textbf{99.1} & \textbf{98.8} & \textbf{67.6} &\textbf{ 66.7} & \textbf{96.5} & \textbf{89.7} \\
\bottomrule
\end{tabular}
}
\end{table*}
As shown in Table~\ref{tab:input_size}, the performance consistently improves as the input resolution increases, with more pronounced gains observed in pixel-level metrics. This suggests that higher-resolution inputs provide richer structural details, which are beneficial for more precise anomaly localization. However, such performance improvements also incur additional computational overhead, including higher memory consumption and longer inference time. Therefore, we adopt an input resolution of $448\times448$, which achieves a favorable balance between detection accuracy and computational cost.

\subsubsection{Ablation on Feature Extractors}
\begin{table*}[t]
\centering
\caption{Effect of feature extractors on anomaly detection and localization performance on MVTec-AD. The evaluated backbones include ResNet, EfficientNet, ViT, DINOv1, and DINOv2.}
\label{tab:backbone}
\setlength{\tabcolsep}{3pt}
\renewcommand{\arraystretch}{0.95}
\footnotesize
\resizebox{0.95\columnwidth}{!}{
\begin{tabular}{c ccc cccc c}
\toprule
\multirow{2}{*}{Backbone} &
\multicolumn{3}{c}{Image-level} &
\multicolumn{4}{c}{Pixel-level} &
\multirow{2}{*}{mAD} \\
\cmidrule(lr){2-4}
\cmidrule(lr){5-8}
& AUROC & AP & F$_1$-max &
AUROC & AP & F$_1$-max & AUPRO & \\
\midrule
ResNet18~\cite{he2016deep} & 94.2 & 98.4 & 95.7 & 96.8 & 50.6 & 54.7 & 91.8 & 83.2 \\
ResNet34~\cite{he2016deep} & 93.2 & 98.0 & 95.2 & 97.1 & 51.7 & 55.2 & 92.3 & 83.2 \\
ResNet50~\cite{he2016deep} & 92.6 & 97.7 & 95.1 & 97.1 & 52.0 & 55.8 & 92.2 & 83.2 \\
ResNet101~\cite{he2016deep} & 92.0 & 97.4 & 95.4 & 97.2 & 51.1 & 54.9 & 92.4 & 82.9 \\
EfficientNet-B0~\cite{tan2019efficientnet} & 97.8 & 99.3 & 97.6 & 97.3 & 53.9 & 58.1 & 92.5 & 85.2 \\
EfficientNet-B2~\cite{tan2019efficientnet} & 98.1 & 99.4 & 98.0 & 97.2 & 49.7 & 55.8 & 92.5 & 84.4 \\
EfficientNet-B4~\cite{tan2019efficientnet} & 98.3 & 99.6 & 98.0 & 97.6 & 54.4 & 57.7 & 94.4 & 85.7 \\
EfficientNet-B6~\cite{tan2019efficientnet} & 98.2 & 99.5 & 98.1 & 97.7 & 54.4 & 57.8 & 94.4 & 85.7 \\
ViT-B/16~\cite{dosovitskiy2020image} & 94.1 & 98.2 & 95.9 & 96.8 & 48.6 & 53.2 & 86.9 & 82.0 \\
ViT-L/16~\cite{dosovitskiy2020image} & 94.9 & 98.1 & 95.9 & 97.1 & 49.9 & 54.0 & 87.2 & 82.4 \\
DINOv1-S~\cite{caron2021emerging} & 94.5 & 97.6 & 95.2 & 98.0 & 62.8 & 62.6 & 92.2 & 86.1 \\
DINOv1-B~\cite{caron2021emerging} & 93.8 & 97.1 & 95.6 & 98.1 & 62.7 & 63.3 & 92.8 & 86.2 \\
DINOv2-S~\cite{oquab2023dinov2} & 99.0 & 99.7 & 98.7 & 98.5 & 64.3 & 63.9 & 95.7 & 88.5 \\
DINOv2-B~\cite{oquab2023dinov2}  & \textbf{99.5} & \textbf{99.9} & \textbf{99.1} & \textbf{98.7} & \textbf{65.6} & \textbf{64.8} & \textbf{95.9} & \textbf{89.1} \\
\bottomrule
\end{tabular}
}

\end{table*}

As shown in Table~\ref{tab:backbone}, we further evaluate the generality of the proposed method across different feature extractors. The evaluated backbones include CNN-based models, i.e., ResNet~\cite{he2016deep} and EfficientNet~\cite{tan2019efficientnet}, as well as transformer-based models, i.e., ViT~\cite{dosovitskiy2020image}, DINOv1~\cite{caron2021emerging}, and DINOv2~\cite{oquab2023dinov2,darcet2023vision}. The proposed method achieves consistently strong performance across both CNN-based and transformer-based backbones, demonstrating its robustness to different feature representations. Among all evaluated backbones, DINOv2-B achieves the best overall performance, suggesting that high-quality self-supervised representations are particularly beneficial for anomaly detection and localization.

\subsubsection{Ablation on Layer Selection}
\begin{table*}[t]
\centering
\caption{Effect of different layer combinations for hierarchical feature extraction on anomaly detection and localization performance on MVTec-AD.}
\label{tab:layer}
\renewcommand{\arraystretch}{1.0}
\setlength{\tabcolsep}{5pt}
\resizebox{0.95\columnwidth}{!}{
\begin{tabular}{c ccc cccc c}
\toprule
\multirow{2}{*}{Layers} &
\multicolumn{3}{c}{Image-level} &
\multicolumn{4}{c}{Pixel-level} &
\multirow{2}{*}{mAD} \\
\cmidrule(r){2-4}
\cmidrule(l){5-8}
& AUROC & AP & F$_1$-max &
AUROC & AP & F$_1$-max & AUPRO \\
\midrule
2 - 5   & 97.2 & 98.6 & 96.3 & 98.4 & 69.2 & 67.2 & 94.7 & 88.8 \\
4 - 7   & 98.9 & 99.6 & 98.3 & 98.6 & \textbf{69.4} & \textbf{67.7} & 95.8 & \textbf{89.8} \\
6 - 9   & 99.3 & 99.7 & 98.9 & \textbf{98.7} & 66.2 & 65.8 & \textbf{96.0} & 89.2 \\
6 -11   & \textbf{99.5} & \textbf{99.9} & \textbf{99.1} & \textbf{98.7} & 65.6 & 64.8 & 95.9 & 89.1 \\
8 -11   & 99.1 & 99.8 & 98.7 & 98.3 & 60.6 & 62.3 & 95.3 & 87.7 \\
10-11   & 98.7 & 99.6 & 98.5 & 97.9 & 56.6 & 59.5 & 94.0 & 86.4 \\
\bottomrule
\end{tabular}
}
\end{table*}
As shown in Table~\ref{tab:layer}, varying the selected feature layers reveals a clear trade-off between image-level detection and pixel-level localization. Deeper layers generally contain higher-level semantic information and therefore lead to stronger image-level performance. However, relying more heavily on deeper representations may slightly reduce pixel-level localization accuracy, since fine-grained spatial details are gradually weakened. For example, using shallow-to-intermediate layers, such as layers 4--7, achieves strong pixel-level performance due to richer local details, but yields relatively lower image-level accuracy. In contrast, using layers 6--11 achieves the best image-level performance while maintaining competitive localization results. Since layers 6--11 achieve the best image-level performance while maintaining competitive pixel-level localization accuracy, we adopt them as the default setting.

\subsubsection{Ablation Study on Diffusion Architecture Robustness}

To evaluate whether the proposed prototype-aware conditioning mechanism is robust across different diffusion architectures, we replace the diffusion denoising backbone with three representative designs, including DiT~\cite{peebles2023scalable}, U-Net~\cite{ronneberger2015u,dhariwal2021diffusion}, and U-ViT~\cite{bao2023all}. For a fair comparison, all variants use the same hierarchical DINOv2 features, local/global prototype design, training objective, and inference setting. The only difference lies in the denoising architecture. In each variant, the learned dual-scale prototypes are incorporated into the denoising process through prototype-aware attention, allowing intermediate features to attend to structured normality priors during reconstruction. For U-Net and U-ViT, prototype-aware attention is inserted into the corresponding attention blocks to enable interactions between intermediate features and learned prototypes.

As shown in Table~\ref{tab:diff_arch}, all three architectures achieve strong anomaly detection and localization performance, indicating that the proposed prototype-aware attention can be effectively integrated into different diffusion backbones. Among them, the DiT-based implementation achieves the best overall performance, with an image-level AUROC of 99.5\%, a pixel-level AUROC of 98.7\%, and an mAD of 89.1\%. U-Net and U-ViT also obtain competitive results, with mAD scores of 88.8\% and 88.5\%, respectively. These results suggest that the effectiveness of prototype-aware conditioning is not restricted to a specific diffusion architecture. Instead, the key factor lies in providing structured normality guidance through dual-scale prototypes, which helps stabilize the reconstruction process across different denoising backbones.

\begin{table*}[t]
\centering
\caption{Ablation study on diffusion architecture robustness. All variants adopt the same dual-scale prototype design and prototype-aware attention mechanism.}
\label{tab:diff_arch}
\renewcommand{\arraystretch}{1.0}
\setlength{\tabcolsep}{5pt}
\resizebox{0.95\textwidth}{!}{
\begin{tabular}{c ccc cccc c}
\toprule
\multirow{2}{*}{Architecture} &
\multicolumn{3}{c}{Image-level} &
\multicolumn{4}{c}{Pixel-level} &
\multirow{2}{*}{mAD} \\
\cmidrule(r){2-4}
\cmidrule(l){5-8}
& AUROC & AP & F$_1$-max &
AUROC & AP & F$_1$-max & AUPRO \\
\midrule
U-Net & 99.3 & 99.8 & 98.7 & 98.4 & 64.3 & \textbf{65.1} & \textbf{96.2} & 88.8 \\
U-ViT & 99.3 & 99.7 & 98.7 & 98.4 & 63.3 & 64.8 & 95.3 & 88.5 \\
DiT & \textbf{99.5} & \textbf{99.9} & \textbf{99.1} & \textbf{98.7} & \textbf{65.6} & 64.8 & 95.9 & \textbf{89.1} \\
\bottomrule
\end{tabular}
}
\end{table*}

\section{Conclusion}
In this paper, we investigated the scalability challenge of MUAD in large category spaces. As category diversity increases, existing unified models often suffer from performance degradation due to increasingly heterogeneous and multi-modal normal distributions. To address this issue, we proposed DPDiff-AD, a dual prototype-conditioned diffusion framework that structures the normal representation space with complementary local and global prototypes and performs prototype-guided diffusion reconstruction for normality modeling. Specifically, local prototypes capture fine-grained patch-level structures, while global prototypes regularize holistic feature distributions through optimal transport. These structured prototype priors are further incorporated into the diffusion reconstruction process via prototype-aware attention, guiding feature reconstruction toward normal manifolds and enhancing anomaly discriminability. Extensive experiments on five industrial anomaly detection benchmarks demonstrate the effectiveness and scalability of DPDiff-AD. In particular, on the large-scale Real-IAD Variety benchmark with 160 categories, DPDiff-AD maintains stable performance under increasing category diversity. Overall, this work provides a structured generative modeling perspective for scalable MUAD and offers a promising direction for robust anomaly detection in large-scale industrial scenarios.

\section*{Statements and Declarations}
\paragraph{Funding}
Yaoxuan Feng acknowledges support from the National Natural Science Foundation of China (NSFC) under Grant No. 625B2145. Bo Chen acknowledges support from the NSFC under Grant No. 62576266, the Fundamental Research Funds for the Central Universities under Grant Nos. QTZX24003 and QTZX23018, and the Overseas Expertise Introduction Project for Discipline Innovation (111 Project) under Grant No. B18039. Wenchao Chen acknowledges support from the NSFC under Grant No. 62571396, the Fundamental Research Funds for the Central Universities under Grant No. QTZX26120, the National Radar Signal Processing Laboratory under Grant No. KGJ202401, and the National Key Laboratory of Electromagnetic Space Security under Grant No. JS20260300296.

\paragraph{Competing Interests}
The authors declare that they have no competing interests relevant to the content of this article.

\paragraph{Data Availability}
The datasets used in this study are available from the following publicly accessible resources:
\begin{itemize}
    \item MVTec-AD: \url{https://www.mvtec.com/research-teaching/datasets/mvtec-ad}.
    \item VisA: \url{https://github.com/amazon-science/spot-diff}.
    \item MPDD: \url{https://github.com/stepanje/MPDD}.
    \item Real-IAD: \url{https://realiad4ad.github.io/Real-IAD/}.
    \item Real-IAD Variety: \url{https://realiad4ad.github.io/Real-IAD-Variety/}.
\end{itemize}

\bibliography{reference}
\newpage

\begin{appendices}

\section{Category Breakdown of S1, S2, and S3 in Real-IAD Variety}
\label{append_S1S2S3_category}

To systematically evaluate the impact of category scale on model performance, we partition the Real-IAD Variety dataset into three subsets, namely \textbf{S1}, \textbf{S2}, and \textbf{S3}, containing 30, 60, and 100 categories, respectively. The category selection follows a randomized strategy to mitigate potential biases in color and material distributions. These category partitions and randomization settings were defined by the original authors of Real-IAD, and we confirmed them through direct communication with the authors.

The category compositions of the three subsets are listed below.

\subsection*{S1 Categories (30 Classes)}
2pin\_block\_plug, 3\_adapter, 3pin\_aviation\_connector, 4\_wire\_stepping\_motor, D\_sub\_connector, DVD\_switch, LED\_indicator, PLCC\_socket, VR\_joystick, access\_card, accurate\_detection\_switch, aircraft\_model\_head, angled\_toggle\_switch, audio\_jack\_socket, bag\_buckle, ball\_pin, balun\_transformer, battery, battery\_holder\_connector, battery\_socket\_connector, bend\_connector, blade\_switch, blue\_light\_switch, bluetooth\_module, boost\_converter\_module, bread\_model, brooch\_clasp\_accessory, button\_battery\_holder, button\_motor, button\_switch.

\subsection*{S2 Categories (60 Classes)}
2pin\_block\_plug, 3\_adapter, 3pin\_aviation\_connector, 4\_wire\_stepping\_motor, D\_sub\_connector, DVD\_switch, LED\_indicator, PLCC\_socket, VR\_joystick, access\_card, accurate\_detection\_switch, aircraft\_model\_head, angled\_toggle\_switch, audio\_jack\_socket, bag\_buckle, ball\_pin, balun\_transformer, battery, battery\_holder\_connector, battery\_socket\_connector, bend\_connector, blade\_switch, blue\_light\_switch, bluetooth\_module, boost\_converter\_module, bread\_model, brooch\_clasp\_accessory, button\_battery\_holder, button\_motor, button\_switch, car\_door\_lock\_switch, ceramic\_fuse, ceramic\_wave\_filter, charging\_port, chip\_inductor, circuit\_breaker, circular\_aviation\_connector, common\_mode\_choke, common\_mode\_filter, connector, connector\_housing\_female, console\_switch, crimp\_st\_cable\_mount\_box, dc\_jack, dc\_power\_connector, detection\_switch, duckbill\_circuit\_breaker, earphone\_audio\_unit, effect\_transistor, electronic\_watch\_movement,ethernet\_connector,ferrite\_bead, ffc\_connector\_plug, flow\_control\_valve, flower\_copper\_shape, flower\_velvet\_fabric, fork\_crimp\_terminal, fuse\_cover, fuse\_holder, gear.

\subsection*{S3 Categories (100 Classes)}
2pin\_block\_plug, 3\_adapter, 3pin\_aviation\_connector, 4\_wire\_stepping\_motor, D\_sub\_connector, DVD\_switch, LED\_indicator, PLCC\_socket, VR\_joystick, access\_card, accurate\_detection\_switch, aircraft\_model\_head, angled\_toggle\_switch, audio\_jack\_socket, bag\_buckle, ball\_pin, balun\_transformer, battery, battery\_holder\_connector, battery\_socket\_connector, bend\_connector, blade\_switch, blue\_light\_switch, bluetooth\_module, boost\_converter\_module, bread\_model, brooch\_clasp\_accessory, button\_battery\_holder, button\_motor, button\_switch, car\_door\_lock\_switch, ceramic\_fuse, ceramic\_wave\_filter, charging\_port, chip\_inductor, circuit\_breaker, circular\_aviation\_connector, common\_mode\_choke, common\_mode\_filter, connector, connector\_housing\_female, console\_switch, crimp\_st\_cable\_mount\_box, dc\_jack, dc\_power\_connector, detection\_switch, duckbill\_circuit\_breaker, earphone\_audio\_unit, effect\_transistor, electronic\_watch\_movement, ethernet\_connector, ferrite\_bead, ffc\_connector\_plug, flow\_control\_valve, flower\_copper\_shape, flower\_velvet\_fabric, fork\_crimp\_terminal, fuse\_cover, fuse\_holder, gear, gear\_motor, green\_ring\_filter, hairdryer\_switch, hall\_effect\_sensor, headphone\_jack\_female, headphone\_jack\_socket, hex\_plug, humidity\_sensor, ingot\_buckle, insect\_metal\_parts, inverter\_connector, jam\_jar\_model, joystick\_switch, kfc\_push\_key\_switch, knob\_cap, laser\_diode, lattice\_block\_plug, lego\_pin\_connector\_plate, lego\_propeller, lego\_reel, lego\_technical\_gear, lego\_turbine, lighting\_connector, lilypad\_led, limit\_switch, lithium\_battery\_plug, littel\_fuse, little\_cow\_model, lock, long\_zipper, meteor\_hammer\_arrowhead, miniature\_laser\_module, miniature\_lifting\_motor, miniature\_motor, miniature\_stepper\_motor, mobile\_charging\_connector, model\_steering\_module, monitor\_socket, motor\_bracket, motor\_gear\_reducer.

\section{More Quantitative Results for Each Category}\label{append Each Category}

To provide a more comprehensive quantitative evaluation, we present additional per-category results. For MVTec-AD, VisA, MPDD, and Real-IAD, DPDiff-AD is compared with existing state-of-the-art methods on each category, as shown in Tables~\ref{append_tab_per_mvtec_detect},~\ref{append_tab_per_mvtec_loc},~\ref{append_tab_per_visa_detect},~\ref{append_tab_per_visa_loc},~\ref{append_tab_per_mpdd_detect},~\ref{append_tab_per_mpdd_loc},~\ref{append_tab_per_realiad_detect}, and~\ref{append_tab_per_realiad_loc}. For each dataset, image-level anomaly detection and pixel-level anomaly localization results are reported separately, using AUROC, AP, F$_1$-max, and AUPRO as evaluation metrics.

In addition, Table~\ref{append_tab_realiad_variety_per_category} provides detailed per-category results of DPDiff-AD on the large-scale Real-IAD Variety benchmark. Covering all 160 categories, this table enables a fine-grained assessment of the scalability of our method under large-scale category diversity.

Together, these results provide a more complete understanding of DPDiff-AD across different datasets and category scales. The comparison tables demonstrate the competitiveness of DPDiff-AD against existing methods, while the Real-IAD Variety results further illustrate its behavior on a substantially larger and more diverse category set.
\begin{table*}[htbp]  
		\setlength{\tabcolsep}{1.5mm}
		\centering \vspace{0mm}
		\caption{ Comparison with SoTA methods on MVTec-AD dataset for multi-class anomaly detection
with AU-ROC/AP/F$_1$-max metrics.}
        \vspace{0mm}
		\resizebox{1\hsize}{!}{
\begin{tabular}{c|c|ccccccc|c}
				\toprule
				\multicolumn{2}{c|}{{ Category}} &RD4AD&UniAD&SimpleNet&DeSTSeg&DiAD&HVQ-Trans&MambaAD&Ours\\
				\midrule
				\multirow{10}{*}{\rotatebox{90}{Object}}
				& Bottle
                & 99.6/99.9/98.4 & 99.7/\textbf{100.}/\textbf{100.} & \textbf{100.}/\textbf{100.}/\textbf{100.}
                & 98.7/99.6/96.8 & 99.7/96.5/91.8 & \textbf{100.}/\textbf{100.}/\textbf{100.}
                & \textbf{100.}/\textbf{100.}/\textbf{100.} & \textbf{100.}/\textbf{100.}/\textbf{100.}\\
                & Cable
                & 84.1/89.5/82.5 & 95.2/95.9/88.0 & 97.5/98.5/94.7
                & 89.5/94.6/85.9 & 94.8/98.8/95.2 & 99.0/98.8/95.1
                & 98.8/99.2/95.7 & \textbf{99.7}/\textbf{99.8}/\textbf{98.9}\\
                & Capsule
                & 94.1/96.9/\textbf{96.9} & 86.9/97.8/94.4 & 90.7/97.9/93.5
                & 82.8/95.9/92.6 & 89.0/97.5/95.5 & 95.4/99.2/96.3
                & 94.4/98.7/94.9 & \textbf{98.1}/\textbf{99.5}/96.8\\
                & Hazelnut
                & 60.8/69.8/86.4 & 99.8/\textbf{100.}/99.3 & 99.9/\textbf{100.}/99.3
                & 98.8/99.2/98.6 & 99.5/99.7/97.3 & \textbf{100.}/\textbf{100.}/99.3
                & \textbf{100.}/\textbf{100.}/\textbf{100.} & \textbf{100.}/\textbf{100.}/\textbf{100.}\\
                & MetalNut
                & \textbf{100.}/\textbf{100.}/99.5 & 99.2/99.9/99.5 & 96.9/99.3/96.1
                & 92.9/98.4/92.2 & 99.1/96.0/91.6 & 99.9/99.9/98.9
                & 99.9/\textbf{100.}/99.5 & \textbf{100.}/\textbf{100.}/\textbf{100.}\\
                & Pill
                & 97.5/99.6/96.8 & 93.7/98.7/95.7 & 88.2/97.7/92.5
                & 77.1/94.4/91.7 & 95.7/98.5/94.5 & 95.8/99.2/94.9
                & 97.0/99.5/96.2 & \textbf{99.5}/\textbf{99.9}/\textbf{99.3}\\
\bottomrule
		\end{tabular}}\vspace{-3mm}
		\label{append_tab_per_mvtec_detect}
	\end{table*}

\begin{table*}[htbp]  
		\setlength{\tabcolsep}{1.5mm}
		\centering \vspace{0mm}
		\caption{ Comparison with SoTA methods on MVTec-AD dataset for multi-class anomaly localization
with AU-ROC/AP/F$_1$-max metrics/AU-PRO metrics.}
        \vspace{0mm}
		\resizebox{1\hsize}{!}{
			\begin{tabular}{c|c|ccccccc|c}
				\toprule
				\multicolumn{2}{c|}{{ Category}} &RD4AD&UniAD&SimpleNet&DeSTSeg&DiAD&HVQ-Trans&MambaAD&Ours\\
				\midrule
				\multirow{10}{*}{\rotatebox{90}{Object}}
				& Bottle
                & 97.8/68.2/67.6/94.0 & 98.1/66.0/69.2/93.1 & 97.2/53.8/62.4/89.0
                & 93.3/61.7/56.0/67.5 & 98.4/52.2/54.8/86.6 & 98.3/71.8/70.2/94.6
                & 98.8/79.7/76.7/95.2 & \textbf{99.3}/\textbf{85.1}/\textbf{83.8}/\textbf{96.9}\\
                & Cable
                & 85.1/26.3/33.6/75.1 & 97.3/39.9/45.2/86.1 & 96.7/42.4/51.2/85.4
                & 89.3/37.5/40.5/49.4 & 96.8/50.1/57.8/80.5 & 98.1/52.4/59.0/87.5
                & 95.8/42.2/48.1/90.3 & \textbf{98.9}/\textbf{76.4}/\textbf{70.0}/\textbf{95.2}\\
                & Capsule
                & 98.8/43.4/50.1/94.8 & 98.5/42.7/46.5/92.1 & 98.5/35.4/44.3/84.5
                & 95.8/47.9/48.9/62.1 & 97.1/42.0/45.3/87.2 & 98.8/45.3/49.7/90.7
                & 98.4/43.9/47.7/92.6 & \textbf{99.2}/\textbf{60.4}/\textbf{56.2}/\textbf{97.2}\\
                & Hazelnut
                & 97.9/36.2/51.6/92.7 & 98.1/55.2/56.8/94.1 & 98.4/44.6/51.4/87.4
                & 98.2/65.8/61.6/84.5 & 98.3/79.2/\textbf{80.4}/91.5 & 98.8/62.7/63.2/92.5
                & 99.0/63.6/64.4/95.7 & \textbf{99.6}/\textbf{83.6}/77.1/\textbf{96.8}\\
                & MetalNut
                & 93.8/62.3/65.4/91.9 & 94.8/55.5/66.4/81.8 & \textbf{98.0}/\textbf{83.1}/79.4/85.2
                & 84.2/42.0/22.8/53.0 & 97.3/30.0/38.3/90.6 & 96.3/67.1/75.5/90.9
                & 96.7/74.5/79.1/93.7 & 97.6/75.0/\textbf{82.2}/\textbf{95.4}\\
                & Pill
                & 97.5/63.4/65.2/95.8 & 95.0/44.0/53.9/95.3 & 96.5/\textbf{72.4}/\textbf{67.7}/81.9
                & 96.2/61.7/41.8/27.9 & 95.7/46.0/51.4/89.0 & 97.1/50.1/57.6/94.9
                & 97.4/64.0/66.5/95.7 & \textbf{97.6}/62.4/63.2/\textbf{98.0}\\
\bottomrule
		\end{tabular}}\vspace{-3mm}
		\label{append_tab_per_mvtec_loc}
	\end{table*}

\begin{table*}[htbp]  
		\setlength{\tabcolsep}{1.5mm}
		\centering \vspace{0mm}
		\caption{Comparison with SoTA methods on VisA dataset for multi-class anomaly detection with
AU-ROC/AP/F$_1$-max metrics.}\vspace{0mm}
		\resizebox{1\hsize}{!}{
			\begin{tabular}{c|c|ccccccc|c}
				\toprule
				\multicolumn{2}{c|}{{ Category}} &RD4AD&UniAD&SimpleNet&DeSTSeg&DiAD&HVQ-Trans&MambaAD&Ours\\
				\midrule
				\multirow{4}{*}{\shortstack{ Complex \\[2mm]  structure} }
				& PCB1
                & 96.2/95.5/91.9 & 96.3/95.5/92.8 & 91.6/91.9/86.0
                & 87.6/83.1/83.7 & 88.1/88.7/80.7 & 96.7/93.2/87.7
                & 95.4/93.0/91.6 & \textbf{97.6}/\textbf{97.3}/\textbf{94.2}\\
                & PCB2
                & \textbf{97.8}/\textbf{97.8}/94.2 & 93.5/94.1/87.2 & 92.4/93.3/84.5
                & 86.5/85.8/82.6 & 91.4/91.4/84.7 & 93.4/94.8/88.0
                & 94.2/93.7/89.3 & 96.8/97.1/\textbf{96.6}\\
                & PCB3
                & 96.4/96.2/91.0 & 89.8/90.6/83.3 & 89.1/91.1/82.6
                & 93.7/95.1/87.0 & 86.2/87.6/77.6 & 92.0/87.1/79.5
                & 93.7/94.1/86.7 & \textbf{97.6}/\textbf{98.6}/\textbf{95.3}\\
                & PCB4
                & \textbf{99.9}/\textbf{99.9}/\textbf{99.0} & 99.4/99.4/96.2 & 97.0/97.0/93.5
                & 97.8/97.8/92.7 & 99.6/99.5/97.0 & 99.5/99.0/97.0
                & \textbf{99.9}/\textbf{99.9}/98.5 & 99.5/99.5/98.5\\
                
				\midrule
				\multirow{4}{*}{\shortstack{ Multiple \\[2mm]  instances} }
				& Macaroni 1
                & 75.9/61.5/76.8 & 88.1/86.2/80.0 & 85.9/82.5/73.1
                & 76.6/69.0/71.0 & 85.7/85.2/78.8 & 93.1/84.1/79.8
                & 91.6/89.8/81.6 & \textbf{95.8}/\textbf{97.8}/\textbf{92.9}\\
                & Macaroni 2
                & 88.3/84.5/83.8 & 81.3/81.1/71.9 & 68.3/54.3/59.7
                & 68.9/62.1/67.7 & 62.5/57.4/69.6 & 86.2/84.1/81.5
                & 81.6/78.0/73.8 & \textbf{93.6}/\textbf{96.0}/\textbf{91.0}\\
                & Capsules
                & 82.2/90.4/81.3 & 73.5/85.6/79.1 & 74.1/82.8/74.6
                & 87.1/93.0/84.2 & 58.2/69.0/78.5 & 77.1/83.4/77.5
                & 91.8/95.0/88.8 & \textbf{98.5}/\textbf{99.0}/\textbf{95.4}\\
\bottomrule
		\end{tabular}}\vspace{-2mm}
		\label{append_tab_per_visa_detect}
	\end{table*}

\begin{table*}[htbp]  
		\setlength{\tabcolsep}{1.5mm}
		\centering \vspace{0mm}
		\caption{Comparison with SoTA methods on VisA dataset for multi-class anomaly localization with
AU-ROC/AP/F$_1$-max/AU-PRO metrics.}\vspace{0mm}
		\resizebox{1\hsize}{!}{
			\begin{tabular}{c|c|ccccccc|c}
				\toprule
				\multicolumn{2}{c|}{{ Category}} &RD4AD&UniAD&SimpleNet&DeSTSeg&DiAD&HVQ-Trans&MambaAD&Ours\\
				\midrule
				\multirow{4}{*}{\shortstack{ Complex \\[2mm]  structure} }
				& PCB1
                & 99.4/66.2/62.4/95.8 & 99.4/68.1/63.3/90.5 & 99.2/86.1/78.8/83.6
                & 95.8/46.4/49.0/83.2 & 98.7/49.6/52.8/80.2 & 99.4/63.1/58.8/87.4
                & \textbf{99.8}/77.1/72.4/92.8 & 99.6/\textbf{88.4}/\textbf{79.7}/\textbf{96.0}\\
                & PCB2
                & 98.0/22.3/30.0/90.8 & 98.3/13.0/19.5/86.0 & 96.6/08.9/18.6/85.7
                & 97.3/14.6/28.2/79.9 & 95.2/07.5/16.7/67.0 & 98.0/10.1/18.2/82.0
                & \textbf{98.9}/13.3/23.4/89.6 & \textbf{98.9}/\textbf{25.1}/\textbf{33.9}/\textbf{92.4}\\
                & PCB3
                & 97.9/26.2/35.2/\textbf{93.9} & 98.5/13.4/23.9/86.1 & 97.2/\textbf{31.0}/\textbf{36.1}/85.1
                & 97.7/28.1/33.4/62.4 & 96.7/08.0/18.8/68.9 & 98.3/21.1/23.7/80.5
                & \textbf{99.1}/18.3/27.4/89.1 & \textbf{99.1}/28.1/34.2/90.6\\
                & PCB4
                & 97.8/31.4/37.0/88.7 & 97.6/34.3/39.0/85.2 & 93.9/23.9/32.9/61.1
                & 95.8/53.0/53.2/76.9 & 97.0/17.6/27.2/85.0 & 97.7/21.1/29.8/85.9
                & \textbf{98.6}/47.0/46.9/87.6 & \textbf{98.6}/\textbf{53.4}/\textbf{53.7}/\textbf{93.0}\\
                
				\midrule
				\multirow{4}{*}{\shortstack{ Multiple \\[2mm]  instances} }
				& Macaroni 1
                & 99.4/02.9/06.9/95.3 & 99.4/10.8/20.1/\textbf{95.6} & 98.9/03.5/08.4/92.0
                & 99.1/05.8/13.4/62.4 & 94.1/10.2/16.7/68.5 & 99.4/09.9/19.3/91.2
                & 99.5/17.5/\textbf{27.6}/95.2 & \textbf{99.6}/\textbf{19.4}/\textbf{27.6}/95.0\\
\bottomrule
		\end{tabular}}\vspace{-2mm}
		\label{append_tab_per_visa_loc}
	\end{table*}

\begin{table*}[htbp]  
		\setlength{\tabcolsep}{1.5mm}
		\centering \vspace{0mm}
				\caption{Comparison with SoTA methods on MPDD dataset for multi-class anomaly detection with
AU-ROC/AP/F$_1$-max metrics.}\vspace{0mm}
		\resizebox{1\hsize}{!}{
			\begin{tabular}{c|cccccc|c}
				\toprule
				Normal Indices &RD4AD&UniAD&SimpleNet&DeSTSeg&HVQ-Trans&MambaAD&Ours\\
				\midrule
				Bracket Black
                & 83.7/87.0/85.7 & \textbf{96.3}/\textbf{96.6}/\textbf{96.9} & 78.7/85.3/80.4
                & 92.9/95.0/91.3  & 90.2/94.5/89.4
                & 82.8/88.6/80.7 & 91.8/96.4/90.7\\
                Bracket Brown
                & 83.1/87.0/91.7 & 93.1/95.8/93.5 & 92.4/95.3/93.5
                & 91.4/92.4/96.2  & 89.7/88.5/89.3
                & \textbf{97.2}/\textbf{98.5}/\textbf{97.1} & 91.6/95.3/90.9\\
                Bracket White
                & 85.0/86.9/80.6 & 82.4/80.5/79.3 & 88.4/91.4/83.0
                & 96.3/93.5/94.9  & 78.0/78.4/75.0
                & 93.6/95.1/89.7 & \textbf{99.6}/\textbf{100.}/\textbf{100.}\\
                Connector
                & 99.8/99.5/96.6 & 82.9/65.0/72.7 & 99.3/98.5/96.6
                & 90.7/77.3/87.5  & 89.5/80.1/80.0
                & 98.8/97.9/92.9 & \textbf{100.}/\textbf{100.}/\textbf{100.}\\
                Metal Plate
                & \textbf{100.}/\textbf{100.}/\textbf{100.} & 58.8/79.6/85.0 & \textbf{100.}/\textbf{100.}/\textbf{100.}
                & \textbf{100.}/\textbf{100.}/\textbf{100.} & 97.3/99.9/98.6
                & 99.8/99.9/99.3 & \textbf{100.}/\textbf{100.}/\textbf{100.}\\
                Tubes
                & 90.4/96.2/88.6 & 67.3/81.6/83.3 & 84.6/94.1/84.6
                & 84.0/92.3/87.2  & 74.3/87.3/82.6
                & 62.9/78.6/81.9 & \textbf{99.0}/\textbf{99.5}/\textbf{97.1}\\
                
                \midrule
                Mean
                & 90.3/92.8/90.5 & 80.1/83.2/85.1 & 90.6/94.1/89.7
                & 92.6/91.8/92.8  & 86.5/88.1/85.8
                & 89.2/93.1/90.3 & \textbf{97.0}/\textbf{98.5}/\textbf{96.5}\\

				\bottomrule
		\end{tabular}}\vspace{0mm}
		\label{append_tab_per_mpdd_detect}
	\end{table*}
\begin{table*}[htbp]  
		\setlength{\tabcolsep}{1.5mm}
		\centering \vspace{0mm}
				\caption{Comparison with SoTA methods on MPDD dataset for multi-class anomaly localization with
AU-ROC/AP/F$_1$-max/AU-PRO metrics.}\vspace{0mm}
		\resizebox{1\hsize}{!}{
			\begin{tabular}{c|cccccc|c}
				\toprule
				Normal Indices &RD4AD&UniAD&SimpleNet&DeSTSeg&HVQ-Trans&MambaAD&Ours\\
				\midrule
				Bracket Black
                & 97.6/05.9/12.3/93.3 & 94.8/00.9/02.2/84.5 & 94.9/03.6/09.1/89.3
                & 91.9/06.0/\textbf{14.2}/87.6  & 96.1/01.1/02.7/87.3
                & 94.9/04.6/13.1/89.1 & \textbf{98.4}/\textbf{06.2}/12.9/\textbf{96.0}\\
                Bracket Brown
                & 96.0/14.2/23.2/94.0 & \textbf{98.6}/\textbf{39.3}/\textbf{47.0}/90.7 & 94.9/08.8/17.0/87.4
                & 78.9/05.1/08.8/66.9  & 98.2/32.3/36.4/86.5
                & 97.8/19.6/27.3/92.8 & 96.2/35.3/39.2/\textbf{94.3}\\
                Bracket White
                & 98.9/01.7/05.0/94.4 & 95.9/00.8/02.8/82.0 & 97.8/02.2/05.6/86.4
                & 91.6/00.9/02.2/52.6  & 94.1/00.6/02.6/79.8
                & \textbf{99.3}/\textbf{08.9}/\textbf{18.9}/\textbf{95.0} & 99.1/03.6/10.0/92.4\\
                Connector
                & 99.1/58.3/54.5/97.2 & 97.6/14.0/24.0/92.1 & 98.9/56.0/55.2/96.5
                & 89.1/31.4/38.3/81.2 & 97.9/18.8/29.7/92.7
                & \textbf{99.2}/55.4/53.4/\textbf{97.5} & 98.7/\textbf{60.8}/\textbf{61.2}/95.4\\
                Metal Plate
                & 99.1/92.6/87.4/96.6 & 93.3/50.4/62.8/80.8 & 98.2/88.9/81.3/88.1
                & 96.9/93.8/87.2/91.4  & 96.6/73.7/74.4/86.3
                & 98.5/86.2/84.3/94.7 & \textbf{99.2}/\textbf{96.2}/\textbf{90.5}/\textbf{97.0}\\
                Tubes
                & 98.9/64.7/61.2/95.7 & 92.1/08.3/14.7/72.7 & 97.9/42.4/46.1/92.6
                & 96.5/46.3/46.7/89.9  & 97.0/38.8/42.8/88.9
                & 96.5/26.7/34.5/87.4 & \textbf{99.2}/\textbf{79.1}/\textbf{74.5}/\textbf{97.1}\\
\bottomrule
		\end{tabular}}\vspace{0mm}
		\label{append_tab_per_mpdd_loc}
	\end{table*}

\begin{table*}[htbp]  
		\setlength{\tabcolsep}{1.5mm}
		\centering \vspace{0mm}
				\caption{Comparison with SoTA methods on Real-IAD dataset for multi-class anomaly detection with
AU-ROC/AP/F$_1$-max metrics.}\vspace{0mm}
		\resizebox{1\hsize}{!}{
			\begin{tabular}{c|ccccccc|c}
				\toprule
				Normal Indices &RD4AD&UniAD&SimpleNet&DeSTSeg&DiAD&HVQ-Trans&MambaAD&Ours\\
				\midrule
				audiojack
                & 76.2/63.2/60.8 & 81.4/76.6/64.9 & 58.4/44.2/50.9
                & 81.1/72.6/64.5 & 76.5/54.3/65.7 & 81.0/80.1/74.5
                & 84.2/76.5/67.4 & \textbf{88.6}/\textbf{84.8}/\textbf{76.3}\\
                bottlecap
                & 89.5/86.3/81.0 & 92.5/91.7/81.7 & 54.1/47.6/60.3
                & 78.1/74.6/68.1 & 91.6/\textbf{94.0}/\textbf{87.9} & 89.0/87.6/77.3
                & 92.8/92.0/82.1 & \textbf{94.1}/93.3/84.1\\
                buttonbattery
                & 73.3/78.9/76.1 & 75.9/81.6/76.3 & 52.5/60.5/72.4
                & 86.7/\textbf{89.2}/83.5 & 80.5/71.3/70.6 & 82.2/88.5/78.9
                & 79.8/85.3/77.8 & \textbf{87.8}/88.9/\textbf{83.7}\\
                endcap
                & 79.8/84.0/77.8 & 80.9/86.1/78.0 & 51.6/60.8/72.9
                & 77.9/81.1/77.1 & 85.1/83.4/\textbf{84.8} & 79.7/85.2/79.4
                & 78.0/82.8/77.2 & \textbf{85.9}/\textbf{87.3}/83.4\\
                eraser
                & 90.0/88.7/79.7 & 90.3/89.2/80.2 & 46.4/39.1/55.8
                & 84.6/82.9/71.8 & 80.0/80.0/77.3 & 89.2/\textbf{89.7}/\textbf{81.9}
                & 87.5/86.2/76.1 & \textbf{92.2}/\textbf{89.7}/80.3\\
                firehood
                & 78.3/70.1/64.5 & 80.6/74.8/66.4 & 58.1/41.9/54.4
                & 81.7/72.4/67.7 & 83.3/81.7/80.5 & \textbf{93.1}/\textbf{85.4}/\textbf{83.1}
                & 79.3/72.5/64.8 & 81.8/72.5/67.5\\
                mint
                & 65.8/63.1/64.8 & 67.0/66.6/64.6 & 52.4/50.3/63.7
                & 58.4/55.8/63.7 & \textbf{76.7}/\textbf{76.7}/\textbf{76.0} & 63.0/75.7/75.1
                & 70.1/70.8/65.5 & 73.6/74.3/67.0\\
\bottomrule
		\end{tabular}}\vspace{0mm}
		\label{append_tab_per_realiad_detect}
	\end{table*}
\begin{table*}[htbp]  
		\setlength{\tabcolsep}{1.5mm}
		\centering \vspace{0mm}
				\caption{Comparison with SoTA methods on Real-IAD dataset for multi-class anomaly localization with
AU-ROC/AP/F$_1$-max/AU-PRO metrics.}\vspace{0mm}
		\resizebox{1\hsize}{!}{
			\begin{tabular}{c|ccccccc|c}
				\toprule
				Normal Indices &RD4AD&UniAD&SimpleNet&DeSTSeg&DiAD&HVQ-Trans&MambaAD&Ours\\
				\midrule
				audiojack
                & 96.6/12.8/22.1/79.6 & 97.6/20.0/31.0/83.7 & 74.4/00.9/04.8/38.0
                & 95.5/25.4/31.9/52.6 & 91.6/01.0/03.9/63.3 & 98.5/\textbf{31.9}/\textbf{41.0}/\textbf{88.0}
                & 97.7/21.6/29.5/83.9 & \textbf{99.3}/16.5/28.7/87.5\\
                bottlecap
                & 99.5/18.9/29.9/95.7 & 99.5/19.4/29.6/96.0 & 85.3/02.3/05.7/45.1
                & 94.5/25.3/31.1/25.3 & 94.6/04.9/11.4/73.0 & 98.4/15.6/22.8/90.0
                & 99.7/30.6/34.6/97.2 & \textbf{99.8}/\textbf{33.8}/\textbf{38.4}/\textbf{97.8}\\
                buttonbattery
                & 97.6/33.8/37.8/86.5 & 96.7/28.5/34.4/77.5 & 75.9/03.2/06.6/40.5
                & 98.3/\textbf{63.9}/\textbf{60.4}/36.9 & 84.1/01.4/05.3/66.9 & 99.0/58.1/59.2/85.3
                & 98.1/46.7/49.5/86.2 & \textbf{99.2}/40.5/47.4/\textbf{90.5}\\
                endcap
                & 96.7/12.5/22.5/89.2 & 95.8/08.8/17.4/85.4 & 63.1/00.5/02.8/25.7
                & 89.6/\textbf{14.4}/\textbf{22.7}/29.5 & 81.3/02.0/06.9/38.2 & 95.6/06.0/14.5/84.7
                & 97.0/12.0/19.6/89.4 & \textbf{99.1}/06.5/13.8/\textbf{95.2}\\
                eraser
                & 99.5/30.8/36.7/96.0 & 99.3/24.4/30.9/94.1 & 80.6/02.7/07.1/42.8
                & 95.8/\textbf{52.7}/\textbf{53.9}/46.7 & 91.1/07.7/15.4/67.5 & 99.2/31.6/38.4/92.4
                & 99.2/30.2/38.3/93.7 & \textbf{99.6}/24.8/31.1/\textbf{96.8}\\
                firehood
                & 98.9/27.7/35.2/87.9 & 98.6/23.4/32.2/85.3 & 70.5/00.3/02.2/25.3
                & 97.3/27.1/35.3/34.7 & 91.8/03.2/09.2/66.7 & 98.9/35.0/\textbf{42.7}/\textbf{93.3}
                & 98.7/25.1/31.3/86.3 & \textbf{99.6}/\textbf{35.5}/40.1/92.8\\
\bottomrule
		\end{tabular}}\vspace{0mm}
		\label{append_tab_per_realiad_loc}
	\end{table*}

\begin{table*}[htbp]
    \setlength{\tabcolsep}{1.2mm}
    \centering
    \vspace{0mm}
    \caption{Detailed per-category results of DPDiff-AD on Real-IAD Variety. We report image-level AUROC/AP/F$_1$-max and pixel-level AUROC/AP/F$_1$-max/AUPRO for each category.}
    \vspace{0mm}
    \resizebox{1\hsize}{!}{
    \begin{tabular}{c|c|ccccccc|c|c|ccccccc}
        \toprule
        \multirow{2}{*}{No.}
        & \multirow{2}{*}{Category}
        & \multicolumn{3}{c}{Image-level}
        & \multicolumn{4}{c|}{Pixel-level}
        & \multirow{2}{*}{No.}
        & \multirow{2}{*}{Category}
        & \multicolumn{3}{c}{Image-level}
        & \multicolumn{4}{c}{Pixel-level} \\
        \cmidrule(lr){3-5} \cmidrule(lr){6-9}
        \cmidrule(lr){12-14} \cmidrule(lr){15-18}
        &
        & AUROC & AP & F$_1$-max
        & AUROC & AP & F$_1$-max & AUPRO
        &
        &
        & AUROC & AP & F$_1$-max
        & AUROC & AP & F$_1$-max & AUPRO \\
        \midrule
        1 & 2pin\_block\_plug & 99.2 & 99.9 & 98.6 & 93.8 & 65.7 & 59.2 & 79.9 & 2 & 3\_adapter & 93.7 & 99.1 & 95.7 & 97.4 & 32.1 & 44.8 & 83.6 \\
        3 & 3pin\_aviation\_connector & 87.7 & 98.2 & 93.7 & 97.4 & 66.3 & 61.3 & 84.1 & 4 & 4\_wire\_stepping\_motor & 82.2 & 97.4 & 94.4 & 64.4 & 1.7 & 4.6 & 37.6 \\
        5 & DVD\_switch & 94.4 & 99.1 & 97.8 & 89.0 & 51.8 & 51.2 & 75.4 & 6 & D\_sub\_connector & 98.4 & 99.7 & 98.5 & 96.9 & 56.5 & 55.8 & 87.0 \\
        7 & LED\_indicator & 89.7 & 98.7 & 94.1 & 85.4 & 6.1 & 14.2 & 61.4 & 8 & PLCC\_socket & 99.1 & 99.9 & 98.2 & 98.8 & 71.0 & 62.7 & 89.4 \\
        9 & VR\_joystick & 95.1 & 99.0 & 95.8 & 97.5 & 54.8 & 53.5 & 86.0 & 10 & access\_card & 90.2 & 98.4 & 95.7 & 62.9 & 13.1 & 21.1 & 74.4 \\
        11 & accurate\_detection\_switch & 88.5 & 98.4 & 94.8 & 93.3 & 58.1 & 53.8 & 77.5 & 12 & aircraft\_model\_head & 100.0 & 100.0 & 99.8 & 98.4 & 45.4 & 48.8 & 90.0 \\
\bottomrule
    \end{tabular}}
    \vspace{0mm}
    \label{append_tab_realiad_variety_per_category}
\end{table*}
\clearpage

\section{Qualitative Visualizations of Anomaly Localization}
\label{append_vision}

\begin{figure*}[t!]\vspace{0mm}
		\centering
		\includegraphics[width=125mm]{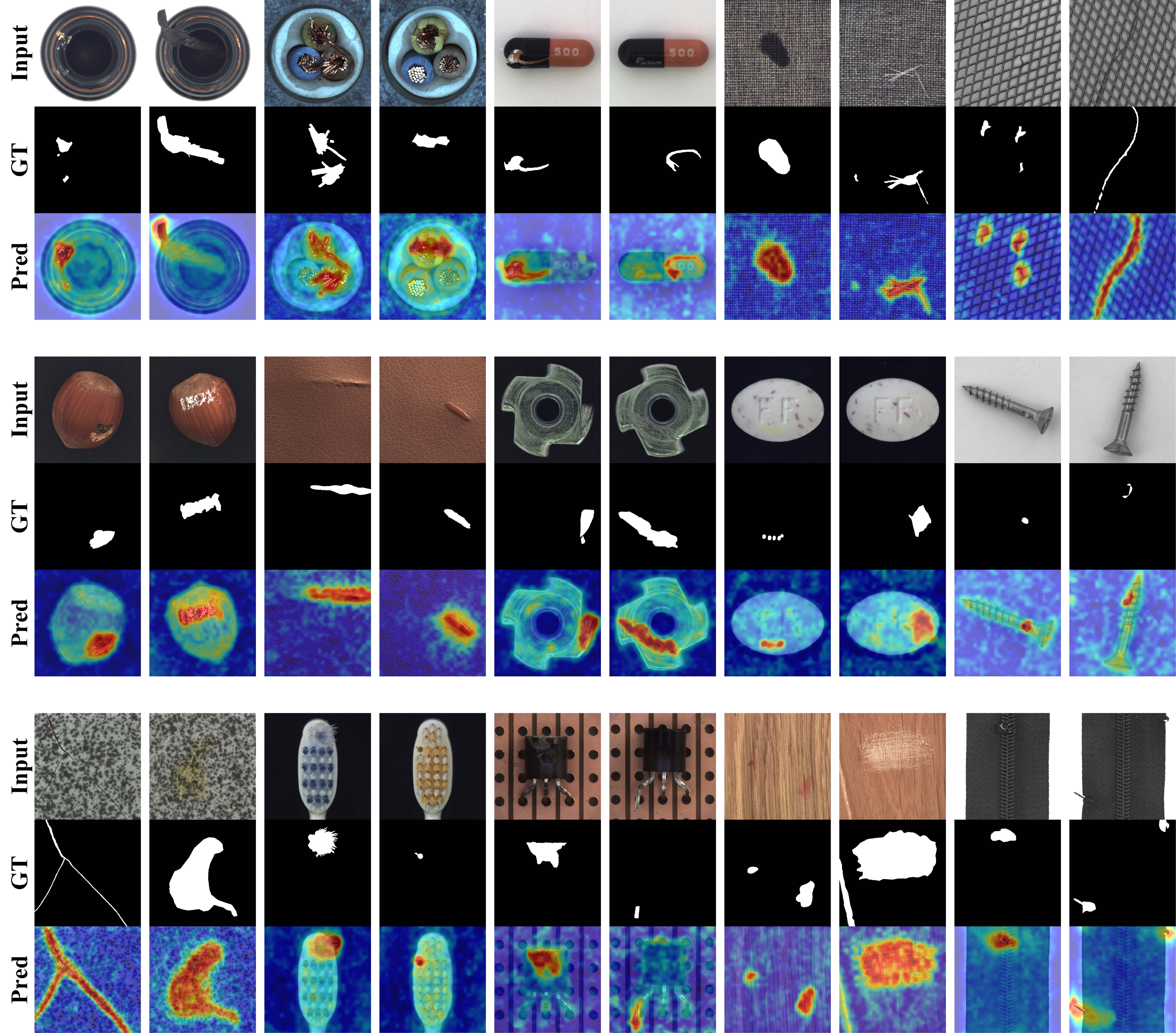} \vspace{0mm}

\caption{Qualitative anomaly localization results of DPDiff-AD on MVTec-AD. From top to bottom, the rows correspond to the input image, ground-truth anomaly mask, and predicted anomaly map.}
        
		 \label{append_mvtec}
		\vspace{0mm}
	\end{figure*}

\begin{figure*}[t!]\vspace{0mm}
		\centering
		\includegraphics[width=125mm]{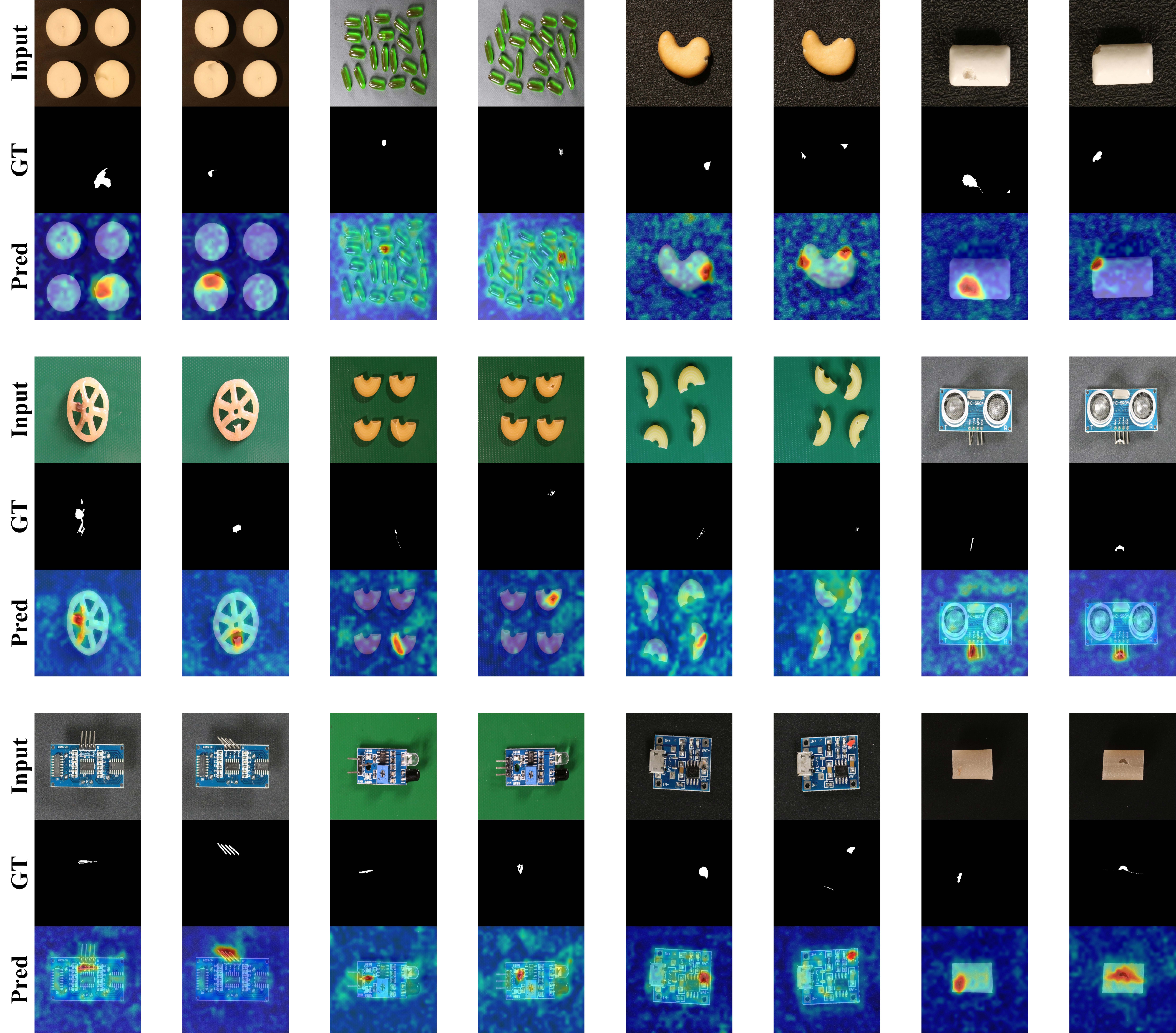} \vspace{0mm}
\caption{Qualitative anomaly localization results of DPDiff-AD on VisA. From top to bottom, the rows correspond to the input image, ground-truth anomaly mask, and predicted anomaly map.}

		 \label{append_visa}
		\vspace{0mm}
	\end{figure*}

\begin{figure*}[t!]\vspace{0mm}
    \centering
    \includegraphics[width=125mm]{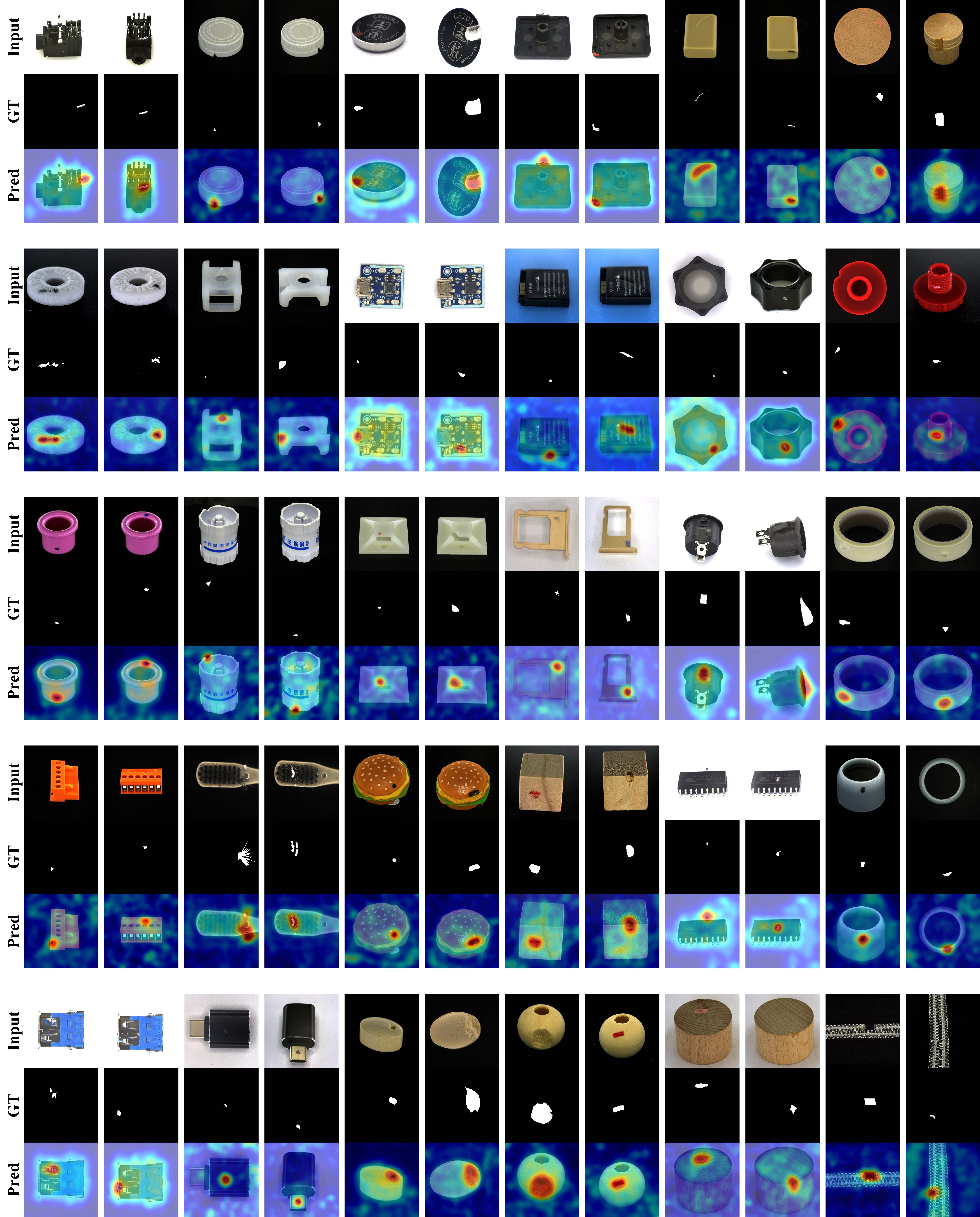} \vspace{0mm}
\caption{Qualitative anomaly localization results of DPDiff-AD on Real-IAD. From top to bottom, the rows correspond to the input image, ground-truth anomaly mask, and predicted anomaly map.}
                    
     \label{append_realiad}
    \vspace{0mm}
\end{figure*}

\begin{figure*}[t!]\vspace{0mm}
    \centering
    \includegraphics[width=125mm]
    {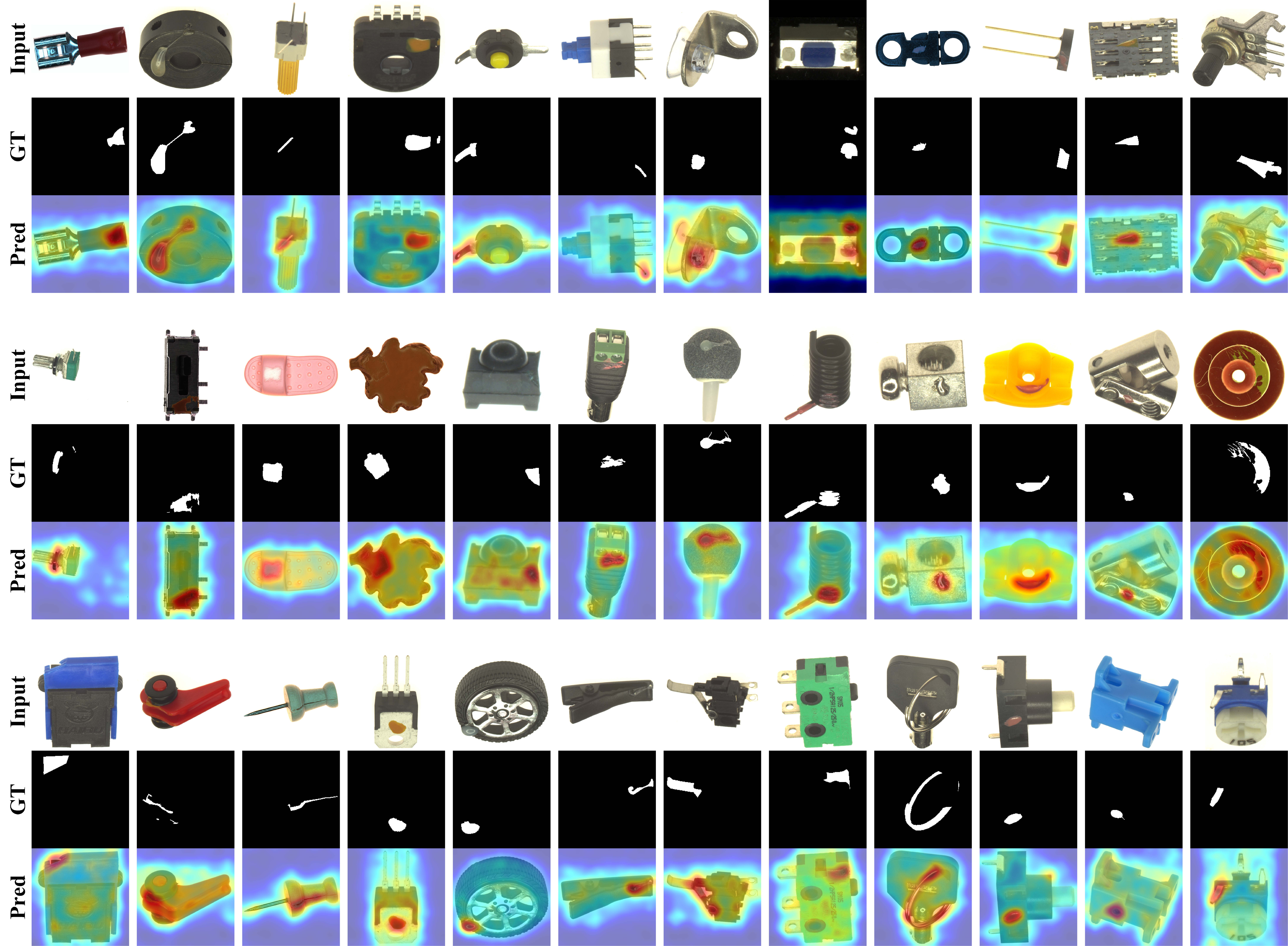} \vspace{0mm}
\caption{Qualitative anomaly localization results of DPDiff-AD on Real-IAD Variety(Part 3). From top to bottom, the rows correspond to the input image, ground-truth anomaly mask, and predicted anomaly map.}
     \label{append_realiad__V_part_3}
    \vspace{0mm}
\end{figure*}

To provide a more comprehensive visual assessment of DPDiff-AD, we present additional anomaly localization visualizations on multiple industrial anomaly detection benchmarks. Specifically, Fig.~\ref{append_mvtec}, Fig.~\ref{append_visa}, and Fig.~\ref{append_realiad} show the visualization results on MVTec-AD, VisA, and Real-IAD, respectively, while Fig.~\ref{append_realiad__V_part_1}, Fig.~\ref{append_realiad__V_part_2}, and Fig.~\ref{append_realiad__V_part_3} present additional results on Real-IAD Variety.

In each visualization, the rows from top to bottom correspond to the input image, ground-truth anomaly mask, and predicted anomaly map. The predicted maps are generally well aligned with the annotated anomalous regions across diverse object and texture categories. Notably, on the large-scale Real-IAD Variety benchmark with 160 categories, DPDiff-AD still produces clear and stable anomaly responses despite substantial category diversity and complex defect variations. These visualizations further demonstrate the effectiveness of DPDiff-AD in localizing anomalies with diverse appearances, scales, and shapes under practical industrial scenarios, while also highlighting its scalability to large-scale category settings.

\begin{figure*}[t!]\vspace{0mm}
    \centering
    \includegraphics[width=125mm]{figure/append_realiad__V_part_1.pdf} \vspace{0mm}
\caption{Qualitative anomaly localization results of DPDiff-AD on Real-IAD Variety(Part 1). From top to bottom, the rows correspond to the input image, ground-truth anomaly mask, and predicted anomaly map.}

     \label{append_realiad__V_part_1}
    \vspace{0mm}
\end{figure*}

\begin{figure*}[t!]\vspace{0mm}
    \centering
    \includegraphics[width=125mm]
    {figure/append_realiad__V_part_2.pdf} \vspace{0mm}
\caption{Qualitative anomaly localization results of DPDiff-AD on Real-IAD Variety(Part 2). From top to bottom, the rows correspond to the input image, ground-truth anomaly mask, and predicted anomaly map.}
     \label{append_realiad__V_part_2}
    \vspace{0mm}
\end{figure*}

\end{appendices}


\end{document}